\documentclass[10pt,journal,compsoc]{IEEEtran}
\usepackage{amsmath,amsfonts}
\usepackage{algorithmic}
\usepackage{array}
\usepackage[caption=false,font=normalsize,labelfont=sf,textfont=sf]{subfig}
\usepackage{textcomp}
\usepackage{stfloats}
\usepackage{url}
\usepackage{verbatim}
\usepackage{graphicx}
\def\BibTeX{{\rm B\kern-.05em{\sc i\kern-.025em b}\kern-.08em
    T\kern-.1667em\lower.7ex\hbox{E}\kern-.125emX}}
\usepackage{balance}
\usepackage{algorithm}
\usepackage{algorithmic}
\makeatletter
\let\c@lofdepth\relax
\let\c@subfigure\relax
\let\l@subfigure\relax
\let\c@subtable\relax
\let\l@subtable\relax

\let\@listsubcaptions\relax
\let\c@subfigure\relax
\let\c@lotdepth\relax
\let\@dottedxxxline\relax
\let\subfloat@label\relax
\let\sf@@sub@label\relax

\makeatother
\usepackage{subfigure}
\usepackage{textcomp,booktabs}
\usepackage[usenames,dvipsnames]{color}
\usepackage{colortbl}
\definecolor{mygray}{gray}{.9}
\usepackage{footnote}

\ifCLASSOPTIONcompsoc
  \usepackage[nocompress]{cite}
\else
  \usepackage{cite}
\fi
\ifCLASSINFOpdf
\else
\fi
\begin{document}
%
\title{Molecular Joint Representation Learning via Multi-modal Information}

\author{Tianyu Wu, Yang Tang,~\IEEEmembership{Senior Member,~IEEE}, Qiyu Sun, Luolin Xiong
\IEEEcompsocitemizethanks{
\IEEEcompsocthanksitem This work was supported in part by the National Natural Science Foundation of China (Basic Science Center Program) under Grant 61988101, in part by the National Natural Science Foundation of China for Distinguished Young Scholars under Grant 61725301 and Grant 61925305. (\emph{Corresponding author: Yang Tang}).
\IEEEcompsocthanksitem Tianyu Wu, Yang Tang, Qiyu Sun and Luolin Xiong are with the Key Laboratory of Smart Manufacturing in Energy Chemical Process, Ministry of Education, East China University of Science and Technology, Shanghai 200237, China (e-mail: tangtany@gmail.com).\protect\\}

}

\markboth{IEEE/ACM Transactions on Computational Biology and Bioinformatics}%
{Shell \MakeLowercase{\textit{et al.}}: Bare Demo of IEEEtran.cls for Computer Society Journals}

\IEEEtitleabstractindextext{%
\begin{abstract}
In recent years, artificial intelligence has played an important role on accelerating the whole process of drug discovery. Various of molecular representation schemes of different modals (e.g. textual sequence or graph) are developed. By digitally encoding them, different chemical information can be learned through corresponding network structures. Molecular graphs and Simplified Molecular Input Line Entry System (SMILES) are popular means for molecular representation learning in current. Previous works have done attempts by combining both of them to solve the problem of specific information loss in single-modal representation on various tasks. To further fusing such multi-modal imformation, the correspondence between learned chemical feature from different representation should be considered. To realize this, we propose a novel framework of molecular joint representation learning via \textbf{M}ulti-\textbf{M}odal information of \textbf{S}MILES and molecular \textbf{G}raphs, called MMSG. We improve the self-attention mechanism by introducing bond-level graph representation as attention bias in Transformer to reinforce feature correspondence between multi-modal information. We further propose a Bidirectional Message Communication Graph Neural Network (BMC GNN) to strengthen the information flow aggregated from graphs for further combination. Numerous experiments on public property prediction datasets have demonstrated the effectiveness of our model.
\end{abstract}

}

\maketitle

\IEEEdisplaynontitleabstractindextext

%
\IEEEpeerreviewmaketitle

\IEEEraisesectionheading{\section{Introduction}\label{sec:introduction}}

%
%
%
%
\IEEEPARstart{I}{n} the traditional drug discovery pipeline, designing an entirely new drug from scratch can cost more than one billion USD and take more than 10 years on average \cite{dimasi2016innovation}. Recently, various artificial intelligence techniques (e.g. Deep learning, Reinforcement learning, etc.) have been applied \cite{9834316,8880542}, aiming at accelerating the entire process \cite{jumper2021highly,jimenez2020drug,9387544}. One crucial step in drug discovery is the construction of the quantitative structure-activity relationship (QSAR) model. Early QSAR models are generally based on fixed molecular representations, e.g. Extended Connectivity Fingerprints \cite{rogers2010extended} and chemical fingerprints. However, these representations greatly rely on hand-crafted features, which restricts their usage \cite{tang2020self}. With the development of deep learning, chemical information can be learned through large amounts of data to construct a data-driven QSAR model. Molecular representations which are generally used in deep learning can be mainly divided into two categories according to the input modal of the data: textual-based representation and graph-based representation \cite{guo2020graseq}.

\begin{figure}[t]
\centering
\includegraphics[width=9cm,height=7.71cm]{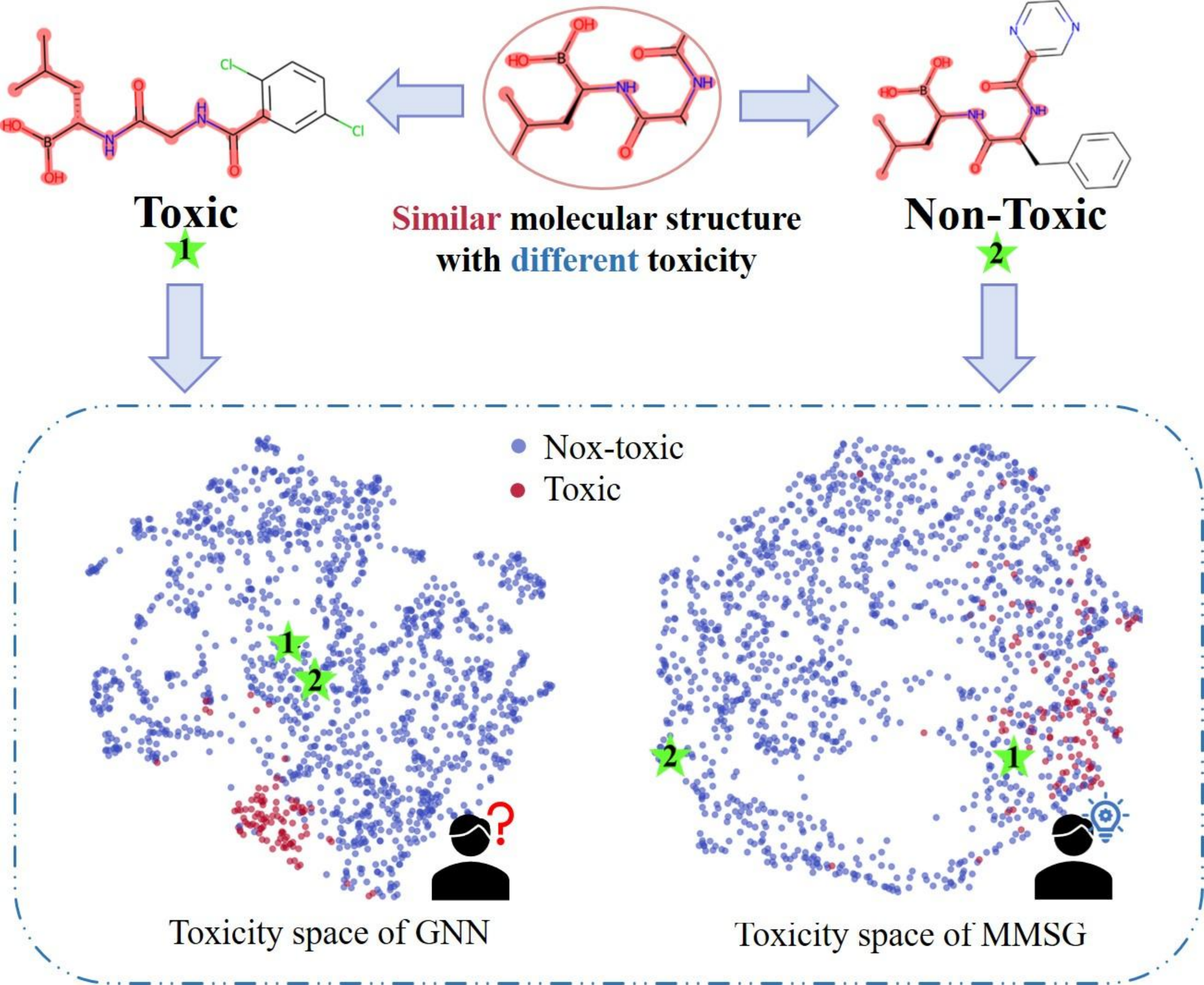}
\caption{Visualization of molecular toxicity space learned by graph-modal input (GNN) and multi-modal input MMSG. Molecules 1 and 2 with different toxicity have similar molecular structure. GNN model (CMPNN here) project them into similar representations due to the alike chemical structure. They are adjacent in corresponding toxicity space and cause problematic final predictions (both non-toxic). By our MMSG framework, multi-modal information are fused to learn more comprehensive chemical feature leading to accurate predictions (one toxic and one non-toxic).} 
\label{fig:1}  
\end{figure}

Simplified Molecular Input Line Entry System (SMILES) \cite{weininger1988smiles} is a kind of textual-based representation with specific syntax rules, which has been widely applied for its simplicity \cite{jimenez2020drug}. For example, in the SMILES string “c1ccccc1”, each lowercase “c” denotes a carbon atom, and “1” represents the start and end of the ring. Thus, architectures borrowed from the field of nature language processing, for example, BERT and Transformer \cite{vaswani2017attention}, can be applied \cite{9866850}. Although SMILES representation can uniquely extract global information of molecules, the loss of spatial chemical structure information still limits the performance of SMILES-based methods \cite{song2020communicative}. 

Unlike SMILES which uses syntax rules to encode molecules, molecular graphs are more natural representations of molecular topology. With the quick development on Graph Neural Network (GNN) \cite{9864145}, researchers are intuitively inspired to utilize molecular graphs for molecular representation learning. Numerous graph-based methods have been gradually proposed \cite{sun2021mocl,li2022geomgcl} with surpassing the performance of SMILES-based methods on various downstream tasks. Despite these abundant progresses, most GNN-based methods still face the problem of losing long-range dependencies and high-order properties with a limited receptive field of GNN \cite{zhu2020beyond}. This limitation may hurt the performance for molecular representation learning which needs whole-graph understanding. A case in Fig. 1 shows that molecules with alike structures may be projected into adjacent space in corresponding functional space by GNN model, causing problematic final predictions, which restricts their usage in real scenarios.

To alleviate the information loss in single-modal representation, an intuitive solution is to incorporate the information of multi-modal representations. Generally, SMILES representation can obvious extract the order of all chemical components with fully considering the long-range contextual information, while graph representation mainly catches the spatial structure information of molecules. Thus, to completely extract these multiple chemical information flow, a multi-modal information processing framework, with complex network structures, is needed. Some previous works have tried combining multi-modal chemical information for various downstream tasks, e.g. retrosynthesis prediction \cite{mao2021molecular} and property prediction \cite{paul2018chemixnet,guo2020graseq}. However, these works mainly focus on improving fusion strategies or network architectures for different tasks instead of considering the correspondence between chemical information embedded in multi-modal representations with completely different encoding rules and the connections between them. According to the syntax rules of SMILES \cite{weininger1988smiles}, most single bonds and aromatic bonds are omitted due to the high usage frequency when representing a molecule which means bond-level infromation are difficlut to learn. While all bonds are deterministically defined in graph-modal representation. This dissymmetry may lead to the information deviation when fusing the features learned by SMILES and graphs. Thus, a motivation is that realizing the correspondence between bond-level chemical information from SMILES and graphs helps to improve the integrality of chemical information for better performance.

Motivated by the above discussions, we propose a novel framework of molecular joint representation learning via \textbf{M}ulti-\textbf{M}odal information of \textbf{S}MILES and molecular \textbf{G}raphs, called MMSG. Focusing on realizing the chemical information correspondence between SMILES and graphs, we make specific improvement on both SMILES and graph pipelines. Detailedly, bond-level representation from graphs is introduced as the self-attention bias in the Transformer for SMILES to better learn the hidden bond information, helping to fuse SMILES and graphs better. We further propose a Bidirectional Message Communication Graph Neural Network (BMC GNN) to strengthen bond and atom representations aggregated from graphs for further combination. Numerous experiments on different public property prediction datasets are conducted, reaching state-of-the-art results. The main contributions of our work are listed as follows:  
\begin{itemize}
\item Different from previous attempts on improving fusing strategies \cite{paul2018chemixnet,mao2021molecular,guo2020graseq}, our MMSG focuses on realizing the chemical information correspondence between SMILES and graphs by introducing bond-level representation, which is omitted in SMILES, as self-attention bias in Transformer to better fuse multi-modal information.
\item To strengthen the information flow aggregated from graphs for further combination, we propose a BMC GNN to update the atom-level and bond-level representations.
\item Numerous experiments on public property prediction datasets are conducted, with achieving 5.8$\%$ relative improvement compared with current state-of-the-art property prediction baselines.
\end{itemize}

 
\section{Related work}
In this section, we briefly review some related works on SMILES-based and graph-based molecular representation learning methods.

\begin{figure*}[h]
\centering
\includegraphics[width=16.5cm,height=10.289cm]{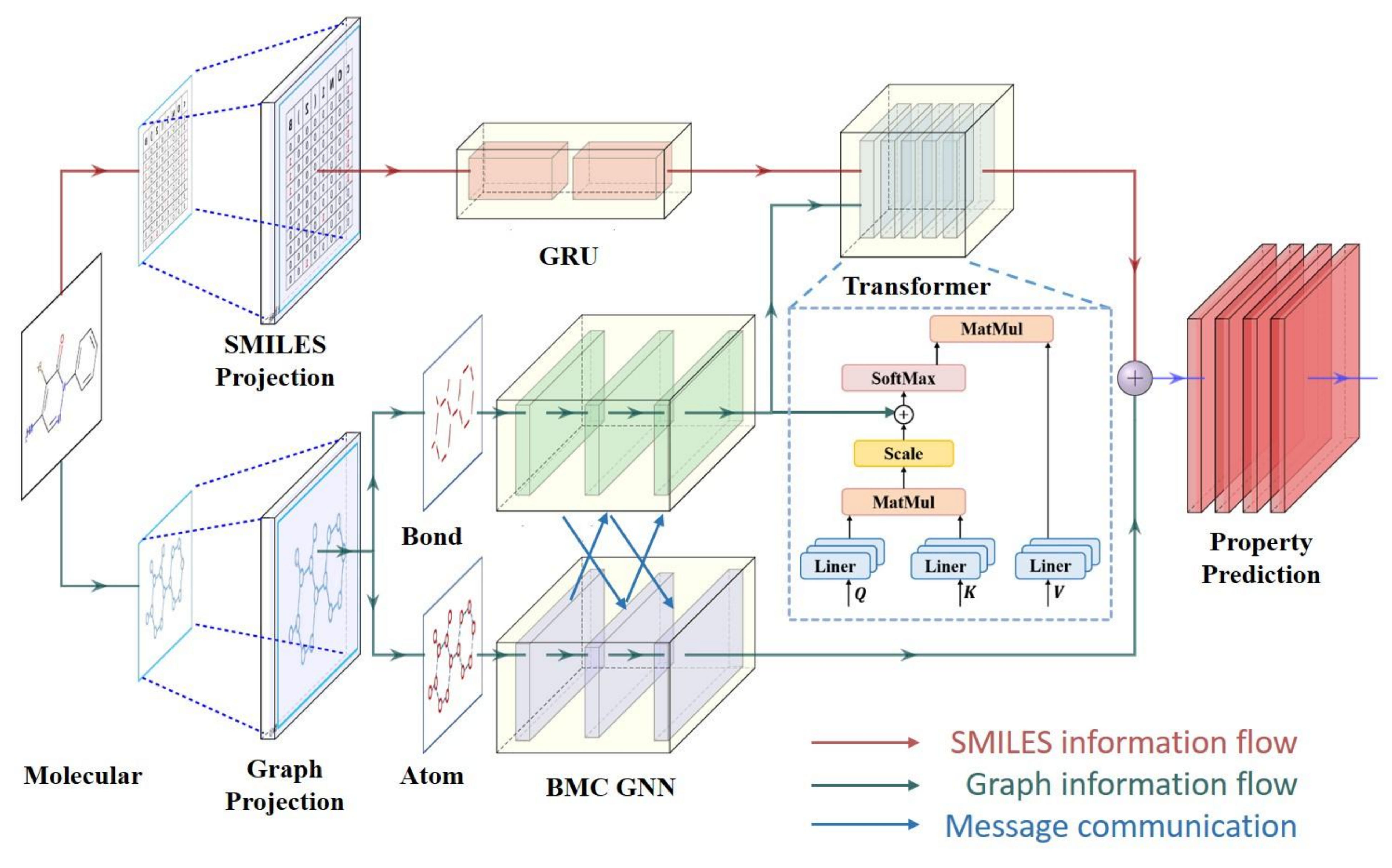} 
\caption{Overview of our MMSG framework. 1) The SMILES representation and the graph representation of a molecule are projected into corresponding matrices. 2) The upper stack consisting of gated recurrent unit (GRU) units and Transformer is used to extract the information from SMILES, and the underlying stack is set for graphs. 3) Bond representation is introduced into Transformer as attention bias, and a single attention layer is shown in detail. BMC GNN is further proposed to strengthen the information flow between atoms and bonds. 4) Finally, both of the network embeddings are concentrated to construct comprehensive eventual molecular representation for property prediction.} 
\label{fig:1}  
\end{figure*}

\subsection{SMILES-based molecular representation learning}
Previously, lots of RNN models have been proposed for SMILES-based molecular representation learning \cite{ozturk2020exploring}. In \cite{zheng2019identifying}, Bi-LSTM is improved by self-attention mechanism for better molecular representation. To further improve the performance of the molecular representation based on SMILES, Transformer-related models have been introduced. Both SMILES-Transformer \cite{honda2019smiles} and SMILES-BERT \cite{wang2019smiles} are firstly pre-trained on large-scale of molecules. Then the model pre-trained can be generalized to various downstream tasks through fine-tuning. Recently, the message passing operation on SMILES has also been presented to learn the hidden rules in chemical strings \cite{jo2020message}. Unfortunately, since SMILES representation lacks the capability to express the spatial chemical structure information and the complex internal connectivity between atoms and bonds, graph-based methods are paid more attention for their better performance on molecular representation learning than SMILES-based methods \cite{sun2020graph}.

\subsection{Graph-based molecular representation learning}
Generally, most of the existing GNN models \cite{hamilton2017inductive,xu2018powerful} used for molecular representation learning follow a Message Passing Neural Network (MPNN) framework \cite{gilmer2017neural}. The MPNN framework includes two steps: a message passing phase and a readout phase. To fully use the message embedded in the edges of molecular graphs, some MPNN variants have been proposed, achieving promising results. In Directed Message Passing Neural Network (DMPNN) \cite{yang2019analyzing}, molecular graphs are seen as directed graphs, and the bond message instead of the atom message has been aggregated for final graph-level representation. In Communicative Message Passing Neural Network (CMPNN) \cite{song2020communicative}, the aggregated bond message has been sent to a communicative kernel to strengthen the atom representation.

However, local aggregating operations in GNN lead to the suspended animation problem that GNN performance drops dramatically when its depth increases \cite{zhang2019gresnet}, causing the loss of long-range dependencies and high-order properties of molecular graphs. Attempting to alleviate this, Transformer-type GNNs have also been studied to learn the long-range dependencies in graph-structured data. In \cite{rong2020self}, a self-supervised pre-training strategy has been developed for molecular representation learning. In \cite{chen2021learning}, a Transformer framework is proposed, which uses both the node and the bond attributes for communicative message interaction. However, Transformer-type GNNs ignore the inductive bias of local substructures, and thus perform poorly in some tasks, where the topology has an important influence on molecular properties \cite{chen2021learning}.

Our MMSG framework, which differs from Transformer-type GNNs, focuses on extracting the long-range contextual information from SMILES and the spatial structure information from graphs, helping to learn more comprehensive molecular representations. Moreover, our framework considers the chemical information correspondence between information embedded in SMILES and graphs for further fusion, helping to learning a better molecular representation.

\section{Methodology}
The key idea of our MMSG is to deal with multi-modal representations with different information embedded, so as to improve the performance of molecular representation learning. To reinforce the correspondence between multi-modal information, we modify the self-attention module in Transformer by novelly introducing bond-level representations from graphs. To further strengthen the information flow aggregated from atoms and bonds, we propose a BMC GNN to update nodes and edges embeddings. Our MMSG framework is shown in Fig. 2.

\subsection{Preliminary}
The main goal of molecular representation learning is to calculate proper molecular representation $H$ through the representation function $R$, according to the given molecular encoding $M$, and it can be formulated as,  
\begin{equation}
   H=R(M).
\end{equation}
Then, molecular representation $H$ will be applied for specific downstream tasks, such as property prediction in this work. In this work, SMILES and molecular graphs are used as the molecular encoding input.

\textbf{SMILES encoding}. Generally, a SMILES sequence with length $T$ can be denoted as $S_{1:T}$ = ($S_1$, ..., $S_T$), $S_t$ $\in$ $\mathcal{D}$, where $\mathcal{D}$ is the token dictionary used to generate SMILES (e.g. ``C", ``N", ``=", ``+"). By mapping these characters in a SMILES sequence into one-hot encoding, we can obtain the input matrix to learn the corresponding molecular representation. In this work, the token dictionary is generated by all the tokens appeared in the datasets.

\begin{table}[t]
\caption{Different atom features used in graph representation.}
\centering
\begin{tabular}{lcl}
\toprule
Features&Size&Description \\
\midrule
Atom type&100&types of atom (e.g. C, H, O)\\
Degree&6&number of neighbor atoms\\
Formal charge&5&integer electronic charge assigned \\
&&to atom \\
Chirality&4&unspecified, tetrahedral CW/CCW, \\
&&or other\\
Number of Hs&5&number of bonded hydrogen atoms\\
Hybridization&5&sp, sp2, sp3, sp3d, or sp3d2\\
Aromaticity&1&whether this atom is part of an \\ 
&&aromatic system \\
atomic mass&1&mass of the atom, divided by 100\\
\bottomrule
\end{tabular}
\label{table 3}
\end{table}

\begin{table}[t]
\caption{Different bond features used in graph representation.}
\centering
\begin{tabular}{lcc}
\toprule
Features&Size&Description \\
\midrule
Bond type&4&single, double, triple or aromatic \\
stereo&6&none, any, E/Z or cis/trans\\
In ring&1&whether the bond is part of a ring  \\
Conjugated&1&whether the bond is conjugated\\
\bottomrule
\end{tabular}
\label{table 3}
\end{table}

\textbf{Graph encoding}. A molecular can be naturally described as an attributed graph $G$ = ($\mathcal{V}$, $\mathcal{E}$), including a set of $n$ nodes (atoms) $|\mathcal{V}|$ = $n$ and a set of $m$ edges (bonds) $|\mathcal{E}|$ = $m$. We use $h_v$ to represent the hidden state of node $v$, and $e_{vw}$ to represent the edge feature of egde $e$ between node $v, w$. Details about the atoms and bonds attributes used to define each node and edge are listed in TABLE 1 and TABLE 2, following \cite{yang2019analyzing}.

In MPNN framework, the message passing phase runs for totally $T$ steps iteratively and is defined in terms of an \textbf{aggregate} function \cite{gilmer2017neural}. At each time step $t$, node message $m_v^{t+1}$ is aggregated to update the hidden states $h_v^t$ at each node in the graph by the update function $U^t$,

\begin{equation}
    m_v^{t+1}=\textbf{aggregate}(h_v^{t},h_w^{t},e_{vu}), u\in{\mathcal{N}(v)},
\end{equation}
\begin{equation}
    h_v^{t+1}=U^t(h_v^{t},m_v^{t+1}),
\end{equation}
where $\mathcal{N}(v)$ is the set of neighbors of node $v$ in graph $G$. Finally, a \textbf{readout} function \cite{gilmer2017neural} is applied to the set of atom representations in $G$ to get the graph-level molecular representation $H$,
\begin{equation}
    H=\textbf{readout}(\{h_v^{T}, \forall v \in \mathcal{V}\}).
\end{equation}
Moreover, the edge message $m_{e_{vw}}$ and edge hidden state $h_{e_{vw}}$ are considered in DMPNN, and a communicative function is further added in CMPNN.

\begin{figure}[t]
\centering
\includegraphics[width=7.10cm,height=6.6cm]{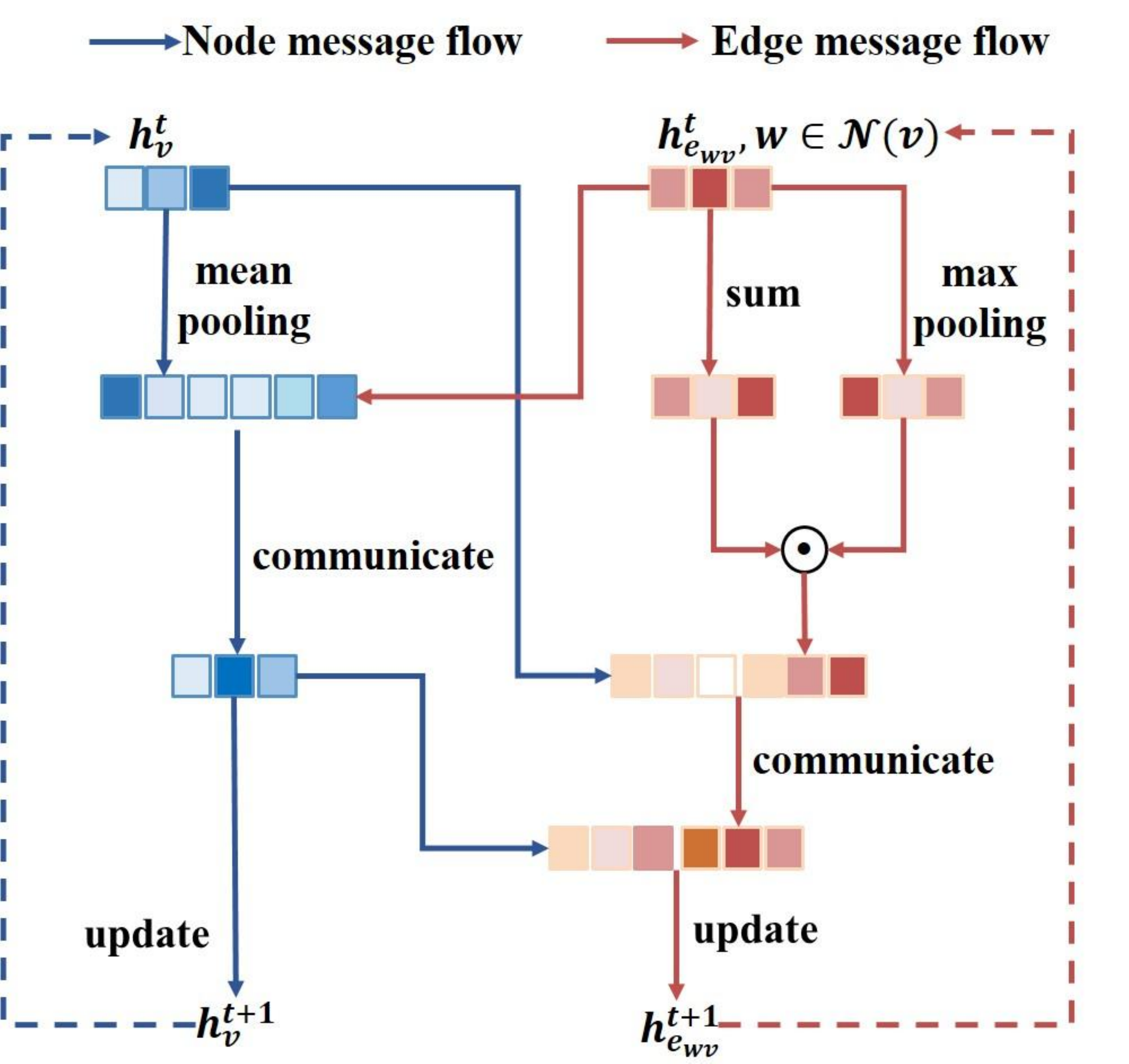}  
\caption{Overview of the message passing procedure of our BMC. Both of node (atom) and edge (bond) message flow are aggregated for molecular representation learning.} 
\label{fig:1}  
\end{figure}

\subsection{Bidirectional Message Communication GNN}
In MPNN, one key step is to leverage the aggregated message to update the hidden state of nodes or edges, referring to $U^t$ in eq (3). However, in the near CMPNN \cite{song2020communicative}, only the node states $h_{e_{vw}}$ are updated iteratively, while the atom states $h_{v}$ are only updated by the communicative kernel which is an addition operator. In MMSG, to reinforce the quality of the aggregated information flow from graphs for further combination, we propose a bidirectional message communication (BMC) GNN to make full use of the node message for more effective message interactions. To realize this, we design different operating functions and specific bi-directional message transportation for nodes and edges. The details are described in Algorithm 1, and the underlined are the modifications compared with CMPNN. 

\begin{algorithm}[tb]
\caption{Bidirectional Message Communication}
\label{alg:algorithm}
\textbf{Input}: Graph $G$=$(\mathcal{V}, \mathcal{E})$; atom and bond attributes\{$x_v$,$\forall v\in\mathcal{V}$; $x_{e_{vw}}$, $\forall e_{vw}\in\mathcal{E}$\};  aggregate functions $\textbf{aggregate}_\mathbf{v}$ and $\textbf{aggregate}_\mathbf{e}$, communicate functions $\textbf{comm}_\mathbf{v}$ and $\textbf{comm}_\mathbf{e}$, update functions $U_v$ and $U_e$, and readout functions $\textbf{readout}_\mathbf{v}$ and $\textbf{readout}_\mathbf{e}$ for nodes and edges respectively; network depth $T$.
\begin{algorithmic}[1] 
\STATE Let $h_v^0 \leftarrow x_v, \forall v \in \mathcal{V}$; $h_{e_{vw}}^0 \leftarrow x_{e_{vw}}, \forall e_{vw} \in \mathcal{E}$ 
\FOR{$t = 0 \cdots T-1 $}
\FOR{$\forall v\in \mathcal{V} $}
\STATE $m_v^{t+1} \leftarrow \textbf{aggregate}_\mathbf{e}($\{$ h_{e_{uv}}^t$, $u\in{\mathcal{N}(v)}$\}$)$
\STATE $p_v^{t+1} \leftarrow \textbf{comm}_\mathbf{v}(m_v^{t+1}, h_v^t)$
\ENDFOR
\FOR{$\forall e\in \mathcal{E} $}
\STATE \underline{$m_{e_{vw}}^{t+1} \leftarrow \textbf{aggregate}_\mathbf{v}(h_v^t,h_w^t)$}
\STATE {\underline{$p_{e_{vw}}^{t+1} \leftarrow \textbf{comm}_\mathbf{e}(m_{e_{vw}}^{t+1},h_{e_{vw}}^t)$}}
\ENDFOR
\FOR{$\forall e\in \mathcal{E} $}
\STATE $p_{e_{vw}}^{t+1} \leftarrow p_v^{t+1}-p_{e_{wv}}^{t+1}$
\STATE $h_{e_{vw}}^{t+1} \leftarrow U_{e}^t(p_{e_{vw}}^{t+1},h_{e_{vw}}^0)$
\ENDFOR
\FOR{$\forall v\in \mathcal{V} $}
\STATE{\underline{$h_v^{t+1} \leftarrow U_{v}^t(p_{v}^{t+1},h_v^0)$}}
\ENDFOR
\ENDFOR
\STATE $m_v^{T} \leftarrow \textbf{aggregate}_\mathbf{e}($\{$ h_{e_{uv}}^{T-1}$,
$u\in{\mathcal{N}(v)}$\}$)$
\STATE $h_v^{T} \leftarrow \textbf{comm}_\mathbf{v}(m_v^{T}, h_v^{T-1})$
\STATE {\underline{$m_{e_{vw}}^{T} \leftarrow \textbf{aggregate}_\mathbf{v}(h_v^{T-1},h_w^{T-1},x_v)$}}
\STATE {\underline{$h_{e_{vw}}^{T} \leftarrow \textbf{comm}_\mathbf{e}(m_{e_{vw}}^{T},h_{e_{vw}}^{T-1},x_{e_{vw}})$}}
\STATE $H_V = \textbf{readout}_\mathbf{v}(\{h_v^{T}, \forall v \in \mathcal{V}\})$
\STATE {\underline{$H_E = \textbf{readout}_\mathbf{e}(\{h_e^{T}, \forall e \in \mathcal{E}\})$}}
\end{algorithmic}
\end{algorithm}

The inputs of the algorithm are the graph $G=(\mathcal{V}, \mathcal{E})$ and all of its atom attributes $x_v$ and bond attributes $x_{e_{vw}}$. Functions are also designed with specific usage. The initial node feature $h_v^0$ is simply the atom attributes, while the initial edge feature $h_{e_{vw}}^0$ is the bond attributes. 

Message aggregation is another crucial step for message passing. Different message aggregators capture different properties of the elements in a molecular graph \cite{xu2018powerful}. The message passing procedure of BMC can be referred to Fig. 3. In our work, we differ the aggregators for node message (line 4) and edge message (line 8). Node message vector $m_v^{t+1}$ is aggregated by the representation its incoming edges in $G$. Edge message vector $m_{e_{vw}}^{t+1}$ is aggregated by the representation of its neighbor nodes.
\begin{equation}
\begin{cases}
\begin{split}
m_v^{t+1}&=\textbf{MAX}(h_{e_{uv}}^{t}) \odot \textbf{SUM}(h_{e_{uv}}^{t}),  u\in{\mathcal{N}(v)},\\
m_{e_{vw}}^{t+1}&=\textbf{MEAN}(h_v^{t},h_w^{t}),
\end{split}
\end{cases}
\end{equation}
where $\mbox{MAX}$ \cite{kipf2016semi}, $\mbox{SUM}$ \cite{xu2018powerful}, $\mbox{MEAN}$ \cite{hamilton2017inductive} are the corresponding aggregating strategy, $\odot$ is an element-wise multiplication operator. 

Then the obtained message vectors of node and edge are  concatenated with the corresponding current hidden states to be sent to the communicate function. For simplicity, we still use an addition operator as the communicative kernel (line 5, 9, 20, 22), and more complex calculating operator can be considered in further research. Considering that the molecular graphs are seen as directed graphs, the hidden state of an edge should not rely on its reverse message which should be subtracted in each updating stage (line 12). 

To further promote the whole message procedure, following \cite{yang2019analyzing}, skip connections to the original feature vector of the node and edge have also been added (line 13, 16). The final update function can be written as,
\begin{equation}
\begin{cases}
\begin{aligned}
h_v^{t+1}&=U_{v}^t(p_{v}^{t+1},h_v^0)=\textbf{ReLU}(h_v^{0}+\mathbf{W_v}\cdot p_v^{t+1}),\\
h_{e_{vw}}^{t+1}&=U_{e}^t(p_{e_{vw}}^{t+1},h_{e_{vw}}^0)=\textbf{ReLU}(h_{e_{vw}}^{0}+\mathbf{W_e}\cdot p_{e_{vw}}^{t+1}),
\end{aligned}
\end{cases}
\end{equation}
where rectified linear unit (\textbf{ReLU}) is the activation function used, and $\mathbf{W_e}, \mathbf{W_v}$ are learned matrices. $p_{e_{vw}}^{t+1}$ and $p_v^{t+1}$ are the message of node and edge after bidirectional communication.

After $T$-1 iteration steps, another iteration is applied, with gathering the initial information, to calculate the set of final node representations $h_v^{T}$ and the bond representations $h_{e_{vw}}^{T}$ in $G$ (line 19-22). 
 
Finally, a readout operation is applied to get the graph-level molecular representation from atoms and bonds. Following the previous work \cite{song2020communicative}, we set different gated recurrent units (GRU) \cite{chung2014empirical} parameterized differently for nodes and edges as readout functions, which can be writted as,
\begin{equation}
    H_V=\sum_{v \in \mathcal{V}}\textbf{GRU}(h_v^{T}),H_E=\sum_{e_{vw} \in \mathcal{E}}\textbf{GRU}(h_{e_{vw}}^{T}),
\end{equation}
Eventually, we gain graph-level molecular representations for the molecule on atoms ${H}_V$ and on bonds ${H}_E$. ${H}_V$ is used to create final molecular representation and ${H}_E$ is used to replenish the bond message of SMILES-based molecular representation in Transformer framework.

\subsection{Transformer-based message self-attention}
Since attention mechanism has shown significant performance on extracting various information \cite{9425008,9456970}, we further propose a Transformer-based message self-attention module to realize the chemical information correspondence between SMILES and graphs. More precisely, we introduce bond-level representation ${H}_E$ calculated by our BMC GNN as the self-attention bias, which helps the model accurately capture the dependency of the hidden bond-level information in SMILES. 

In our Transformer framework, we do not catch the positional encoding \cite{vaswani2017attention} of the SMILES sequence data. Instead, we firstly use a bidirectional GRU unit to pre-process the one-hot encoding to capture the feature vector of the sequence in SMILES, and this progress can be formulated as,
\begin{equation}
\begin{cases}
   \overrightarrow{h_i} = \overrightarrow{\textbf{GRU}}(t_i,\overrightarrow{h_{i-1}}),\\
   \overleftarrow{h_i} = \overleftarrow{\textbf{GRU}}(t_i,\overleftarrow{h_{i-1}}),
\end{cases}
\end{equation}
where $\overrightarrow{h_i},\overleftarrow{h_i}$ are bidirectional hidden states for the $i$ th token of a SMILES string embedded by GRU, and a hidden state $h_i$ is obtained to replace token embedding $t_i$ as,
\begin{equation}
    h_i = (\overrightarrow{h_i},\overleftarrow{h_i}).
\end{equation}
Finally, we use $H_s$ to denote the contextual representation of a SMILES string with length $n$ as,
\begin{equation}
    H_s = (h_0, h_1, \cdots, h_n).
\end{equation}

Then $H_s$ is sent to Transformer to learn long-range information hidden in SMILES. The input is mapped into different embeddings $(Q, K, V)$ with corresponding matrices $W$, where $(Q, K, V)$ represents queries, keys and values. The \textbf{Attention} function \cite{vaswani2017attention} for self-attention in Transformer can be defined as:
\begin{equation}
\begin{cases}
\begin{aligned}
    Q=H_{s}W^Q, K=H_{s}W^K,V=H_{s}W^V,\\
    \textbf{Attention}(H_s)=\textbf{softmax}(QK^\top/\sqrt{d_K})V,
\end{aligned}
\end{cases}
\end{equation}
where $d_K$ is the dimension of $K$ and \textbf{softmax} is a softmax function. We omit the bias term of multi-head attention here for simplicity since it is a straightforward extension of single-head self-attention. 

Considering that in SMILES strings, feature of each atom can be intuitively extracted from word tokens, while most of the single and aromatic bonds are omitted. Following this line, we naturally use bond representations from graphs to reinforce the correspondence on bond-level information learned by SMILES and graphs with the help of attention mechanism, so as to improve the quality of the final molecular representation. We also apply linear transformation for $H_E$ from BMC GNN to deal with the different length of SMILES tokens and bond features, thus they can be processed correctly. The improved self-attention matrix can be written as:
\begin{equation}
\textbf{Attention}(H_s)=\textbf{softmax}(QK^T/\sqrt{d_K}+H_E)V,
\end{equation}

By applying layer normalization (\textbf{LN}) before multi-head self-attention (\textbf{MHA}) blocks and feed-forward network (\textbf{FFN}) blocks \cite{vaswani2017attention}, we get the final sequence representation ${H_{S}}$ as:
\begin{equation}
    H_s^{'}=\textbf{MHA}(\textbf{LN}(H_s))+H_s,
\end{equation}
\begin{equation}
    H_S=\textbf{FFN}(\textbf{LN}(H_s^{'}))+H_s^{'}.
\end{equation}

Finally, by combining the representation $H_S$ and $H_V$ from SMILES and graphs, molecular properties $Y$ can be further predicted through an independent feed-forward network (\textbf{FFN}):
\begin{equation}
    Y=\textbf{FFN}(H_S+H_V).
\end{equation}

\section{Experiments}
\subsection{Benchmark datasets}
To display the molecular representation capability of our MMSG framework, we conduct experiments on 7 public benchmark datasets from MolecularNet \cite{wu2018moleculenet} across classification and regression tasks. TABLE 3 summarizes all the information of the benchmark datasets used, including task types, dataset sizes and the metric used.  Among all of them, BBBP \cite{martins2012bayesian}, Tox21 \cite{tox212017}, SIDER \cite{kuhn2016sider} and ClinTox \cite{gayvert2016data} are used for binary or multi-binary classification tasks, and FreeSolv \cite{mobley2014freesolv}, ESOL \cite{delaney2004esol} and Lipophilicty \cite{wenlock2015experimental} are used for regression tasks. 

Details of each dataset are listed as follows. The Blood–brain barrier penetration (BBBP) dataset \cite{martins2012bayesian} contains compounds on their permeability properties with binary labels. The “Toxicology in the 21st Century” (Tox21) dataset \cite{tox212017} 
contains large amounts of experimental compounds for 12 different targets relevant to drug toxicity. The Side Effect Resource (SIDER) dataset \cite{kuhn2016sider} contains marketed drugs along with side effects and these drugs are grouped into 27 system organ classes. The ClinTox dataset \cite{gayvert2016data} contains drugs with FDA approval status and clinical trial toxicity. The Free Solvation (FreeSolv) dataset \cite{mobley2014freesolv} contains measurements on several compounds with its calculated hydration free energy of in water. ESOL dataset \cite{delaney2004esol} contains water solubility data for molecules. Lipophilicity dataset \cite{wenlock2015experimental} contains the experimental results of octanol/water distribution coefficient of the molecules.

\begin{table}[t]
\caption{Details about the bechmark datasets used, showing task numbers, task types, molecule numbers and the metric used. ``C" means the classification task and ``R" means the regression task. }
\centering
\begin{tabular}{lcccc}
\toprule
Dataset&Tasks&Type&Molecules&Metric \\
\midrule
BBBP&1&C&2040&ROC-AUC\\
Tox21&12&C&7812&ROC-AUC\\
SIDER&27&C&1396&ROC-AUC\\
ClinTox&2&C&1478&ROC-AUC\\
\midrule
FreeSolv&1&R&639&RMSE\\
ESOL&1&R&1127&RMSE\\
Lipophilicity&1&R&4200&RMSE\\
\bottomrule
\end{tabular}
\label{table 3}
\end{table}

\subsection{Baseline models}
We compare MMSG with 16 baseline methods, including the models appeared in the MolecularNet and several state-of-the-art approaches. All of the baseline methods are summarized in TABLE 4. These methods can be mainly divided into different types according to the molecular representation and methods used. RNN \cite{chung2014empirical}, Tranformer \cite{vaswani2017attention} and MPAD \cite{jo2020message} use SMILES sequence as the input with different network structures. GCN \cite{kipf2016semi}, Weave \cite{kearnes2016molecular}, N-Gram \cite{liu2019n} and Attentive FP \cite{xiong2019pushing} are graph convolutional methods. MPNN \cite{gilmer2017neural}, DMPNN \cite{yang2019analyzing}, CMPNN \cite{song2020communicative}, GROVER \cite{rong2020self} and CoMPT \cite{chen2021learning} follow the message passing methods operated on directed graphs. MoCL \cite{sun2021mocl} introduces contrastive learning, while HamNet \cite{li2021hamnet} and GeomGCL \cite{li2022geomgcl} incorporate 3D geometry of molecules. GraSeq \cite{guo2020graseq} and our MMSG mainly focus on combing multi modal molecular representations to improve the final performance. In our work, we choose more graph-based methods for comparison so that to prove the effectiveness of combing multi-modal representations.

\begin{table}[t]
\caption{Summary of all baseline methods used.}
\centering
\begin{tabular}{lcc}
\toprule
 Methods&  Citation& Representation  \\
\midrule
RNN(GRU)&\cite{chung2014empirical}&SMILES\\
Transformer&\cite{vaswani2017attention}&SMILES\\
MPAD &\cite{jo2020message}&SMILES\\
GCN&\cite{kipf2016semi}&Graph \\
Weave&\cite{kearnes2016molecular}&Graph\\
N-Gram&\cite{liu2019n}&Graph \\
Attentive FP &\cite{xiong2019pushing}&Graph \\ 
MPNN &\cite{gilmer2017neural}&Graph\\
DMPNN &\cite{yang2019analyzing}&Graph\\
CMPNN (IJCAI 20) &\cite{song2020communicative}&Graph\\
GROVER (NeruIPS 20) &\cite{rong2020self}&Graph\\
CoMPT (IJCAI 21)&\cite{chen2021learning}&Graph\\
HamNet (ICLR 21) &\cite{li2021hamnet}&Graph\\
MoCL (KDD 21)&\cite{sun2021mocl}&Graph\\
GeomGCL (AAAI 22)& \cite{li2022geomgcl}&Graph\\
GraSeq (CIKM 20)&\cite{guo2020graseq}&Graph/SMILES\\
MMSG (Ours)&$-$&Graph/SMILES\\
\bottomrule
\end{tabular}
\label{table 3}
\end{table}

\begin{table*}[h]
\caption{Summary of the hyper-parameter ranges used in MMSG.}
\centering
\begin{tabular}{lll}
\toprule
Hyper-Parameters& Descriptions&Range \\
\midrule
Batch size& batch size used during the training of the model&50, 64, 128 \\
Warmup Epochs&number of epochs during which the learning rate changes &5, 10\\
Epoch&number of epochs for training&50, 100\\
Layer number (GNN)&the number of Bidirectional Message Communication GNN layers&2, 5\\
Layer number (GRU)&the number of GRU unit&3\\
Layer number (FFN)&the number of Feed-Forward Network layers&2, 5\\
Layer number (Trans)&the number of Transformer layers& 6, 12\\
Attention heads number&number of self-attention heads in Transformer& 16, 32\\
GNN Hidden&dimensionality of hidden layers in GNN& 128, 256, 300\\
GRU Hidden&dimensionality of hidden layers in GRU& 128, 256, 300\\
FNN Hidden&dimensionality of hidden layers in FNN& 128, 256, 300\\
Transformer Hidden&dimensionality of hidden layers in Transformer& 128, 256, 300\\
Initial Learning rate&initial learning rate of Noam learning rate scheduler& 1e-3, 1e-4, 1e-5\\
Max Learning rate&max learning rate of Noam learning rate scheduler& 2e-3, 2e-4, 2e-5\\
Final Learning rate&learning rate of Noam learning rate scheduler& 1e-3, 1e-4, 1e-5\\
\bottomrule
\end{tabular}
\label{table 3}
\end{table*}

\begin{table*}[h]
\caption{Quantitative results on scaffold splitting of different baseline models and our MMSG, and ``Trans" means the Transformer model. Mean and standard deviation of ROC-AUC or RMSE values are reported. For different metrics, ``$\uparrow$" means the higher the better, and ``$\downarrow$" contrarily. Results in bold and cells colored gray denote the best and the second best, respectively.}
\centering
\begin{tabular}{lcccc|ccc}
\toprule
Task & \multicolumn{4}{c|}{Classification (ROC-AUC$\uparrow$ Higher is better)} & \multicolumn{3}{c}{Regression  (RMSE$\downarrow$ Lower is better)}\\
\midrule
Dataset &  BBBP & Tox21 & SIDER & ClinTox & FreeSolv & ESOL & Lipophilicity\\
\midrule
RNN&0.832 $\pm$ 0.015&0.725 $\pm$ 0.017&0.593 $\pm$ 0.010&0.879 $\pm$ 0.009&2.232 $\pm$ 0.299&0.950 $\pm$ 0.034&0.891 $\pm$ 0.009\\
Trans&0.900 $\pm$ 0.053&0.706 $\pm$ 0.021&0.559 $\pm$ 0.017&0.905 $\pm$ 0.064&2.246 $\pm$ 0.237&1.144 $\pm$ 0.118&1.169 $\pm$ 0.031\\
\midrule
GCN&0.877 $\pm$ 0.036&0.772 $\pm$ 0.041&0.593 $\pm$ 0.035&0.845 $\pm$ 0.051&2.900 $\pm$ 0.135&1.068 $\pm$ 0.050&0.712 $\pm$ 0.049\\
Weave&0.837 $\pm$ 0.065&0.741 $\pm$ 0.044&0.543 $\pm$ 0.034&0.823 $\pm$ 0.023&2.398 $\pm$ 0.250&1.158 $\pm$ 0.055&0.813 $\pm$ 0.042\\
N-Gram&0.912 $\pm$ 0.013&0.769 $\pm$ 0.027&0.632 $\pm$ 0.005&0.855 $\pm$ 0.037&2.512 $\pm$ 0.190&1.100 $\pm$ 0.160&0.876 $\pm$ 0.033\\
AttenFP&0.908 $\pm$ 0.050&0.807 $\pm$ 0.020&0.605 $\pm$ 0.060&0.933 $\pm$ 0.020&2.030 $\pm$ 0.420&0.853 $\pm$ 0.060&0.650 $\pm$ 0.030\\
\midrule
MPNN&0.913 $\pm$ 0.041&0.808 $\pm$ 0.024&0.595 $\pm$ 0.030&0.879 $\pm$ 0.054&2.185 $\pm$ 0.952&1.167 $\pm$ 0.430&0.672 $\pm$ 0.051\\
DMPNN&0.919 $\pm$ 0.030&{\cellcolor{mygray}{0.826 $\pm$ 0.023}}& 0.632 $\pm$ 0.023&0.897 $\pm$ 0.040&2.177 $\pm$ 0.914&0.980 $\pm$ 0.258&0.653 $\pm$ 0.046\\
CMPNN&0.927 $\pm$ 0.017&0.806 $\pm$ 0.016&0.616 $\pm$ 0.003&0.902 $\pm$ 0.008 &2.007 $\pm$ 0.442&0.798 $\pm$ 0.112&0.614 $\pm$ 0.029\\
GROVER&0.911 $\pm$ 0.008&0.803 $\pm$ 0.020&0.624 $\pm$ 0.006&0.884 $\pm$ 0.013&1.987 $\pm$ 0.072&0.911 $\pm$ 0.116&0.643 $\pm$ 0.030\\
CoMPT&{\cellcolor{mygray}{0.938 $\pm $0.021}}&0.809 $\pm$ 0.014&{\cellcolor{mygray}{0.634 $\pm$ 0.030}}& {\cellcolor{mygray}{0.934 $\pm$ 0.019}}& {\cellcolor{mygray}{1.855 $\pm$ 0.578}}&{\cellcolor{mygray}{0.774 $\pm$ 0.058}}&{\cellcolor{mygray}{0.592 $\pm$ 0.048}}\\
\midrule
GraSeq&0.926 $\pm$ 0.011&0.755 $\pm$ 0.004&0.609 $\pm$ 0.009&0.904 $\pm$ 0.004&2.218 $\pm$ 0.184&1.034 $\pm$ 0.035&0.780 $\pm$ 0.012\\
MMSG (ours)&\textbf{0.957 $\pm$ 0.019}&\textbf{0.834 $\pm$ 0.008}&\textbf{0.655 $\pm$ 0.022}&\textbf{0.992 $\pm$ 0.005}&\textbf{1.638 $\pm$ 0.525}&\textbf{0.688 $\pm$ 0.030} &\textbf{0.577 $\pm$ 0.014}\\ 
\bottomrule
\end{tabular}
\label{table 1}
\end{table*}

\begin{table*}[h]
\caption{Quantitative results on random splitting of the state-of-the-art models and our MMSG, and ``Trans" means the Transformer model. Mean and standard deviation of AUC or RMSE values are reported. For different metrics, ``$\uparrow$" means the higher the better, and ``$\downarrow$" contrarily. ``$-$" means the results are not reported. Results in cells bolded and cells colored gray denote the best and the second best, respectively.}
\centering
\begin{tabular}{lcccc|ccc}
\toprule
Task & \multicolumn{4}{c|}{Classification (ROC-AUC$\uparrow$) Higher is better} & \multicolumn{3}{c}{Regression  (RMSE$\downarrow$ Lower is better)}\\
\midrule
Dataset &  BBBP & Tox21 & SIDER & ClinTox & FreeSolv & ESOL & Lipophilicity\\
\midrule
RNN&0.902 $\pm$ 0.010&0.806 $\pm$ 0.007&0.605 $\pm$ 0.009&0.915 $\pm$ 0.009&1.108 $\pm$ 0.146&0.743 $\pm$ 0.020&0.770 $\pm$ 0.025\\
Trans&0.944 $\pm$ 0.011&0.813 $\pm$ 0.013&0.602 $\pm$ 0.012&0.954 $\pm$ 0.003&1.021 $\pm$ 0.102&0.767 $\pm$ 0.079&0.909 $\pm$ 0.023\\
MPAD&0.942&0.855&0.637&\cellcolor{mygray}{0.987}&$-$&$-$&$-$\\
\midrule
GCN&0.690&0.829&0.638&0.807&1.40&0.970&-\\
Weave&0.671&0.820&0.581&0.832&1.220&0.610&-\\
N-Gram&0.890&0.842&-&0.870&1.371&0.718&-\\
AttenFP&-&0.858&0.637&0.940&0.736&0.503&0.578\\
\midrule
MPNN&0.910 $\pm$ 0.032&0.844 $\pm$0 .014&0.641 $\pm$ 0.014&0.881 $\pm$ 0.037&1.242 $\pm$ 0.249&0.702 $\pm$ 0.042&0.645 $\pm$ 0.075\\
DMPNN&0.913 $\pm$ 0.026&0.845 $\pm$ 0.015&0.646 $\pm$ 0.016&0.894 $\pm$ 0.027&1.167 $\pm$ 0.150&0.665 $\pm$ 0.052&0.596 $\pm$ 0.050\\
CMPNN&{\cellcolor{mygray}{0.963 $\pm$ 0.003}}&0.856 $\pm$ 0.007&{\cellcolor{mygray}{0.666 $\pm$ 0.007}}&0.933 $\pm$ 0.012&0.819 $\pm$ 0.147&0.547 $\pm$ 0.011&0.560 $\pm$ 0.047\\
GROVER&0.955 $\pm$ 0.003&0.842 $\pm$ 0.009&0.663 $\pm$ 0.008&0.929 $\pm$ 0.178&0.974 $\pm$ 0.069&0.677 $\pm$ 0.017&0.686 $\pm$ 0.006\\
CoMPT&0.953 $\pm$ 0.002&0.849 $\pm$ 0.016&0.657 $\pm$ 0.027&{0.958 $\pm$ 0.020}&0.940 $\pm$ 0.128&0.538 $\pm$ 0.043&{\cellcolor{mygray}{0.554 $\pm$ 0.037}}\\
\midrule
HamNet&$-$&\textbf{0.875 $\pm$ 0.006}&$-$&$-$&{\cellcolor{mygray}{0.731 $\pm$ 0.024}}&{\cellcolor{mygray}{0.504}}$\pm$0.016&0.557 $\pm$ 0.014\\
MoCL&0.911&0.824&0.628&0.750&1.478&0.934&0.742\\
GeomGCL&$-$&0.850&0.648&0.919&0.866&0.555&\cellcolor{mygray}{0.541}\\
\midrule
GraSeq&0.942 $\pm$ 0.012&0.810 $\pm$ 0.008&0.620 $\pm$ 0.008&0.918 $\pm$ 0.005&0.865 $\pm$ 0.032&0.652 $\pm$ 0.039&0.648 $\pm$ 0.041\\
MMSG (ours)&\textbf{0.980 $\pm$ 0.004}&{\cellcolor{mygray}{0.861 $\pm$ 0.031}}&\textbf{0.691 $\pm$ 0.015}&\textbf{0.993 $\pm$ 0.004}&\textbf{0.712 $\pm$ 0.087}&\textbf{0.495 $\pm$ 0.017}&\textbf{0.538 $\pm$ 0.007}\\
\bottomrule
\end{tabular}
\label{table 1}
\end{table*}

\begin{table*}[h]
\caption{Ablation studies on the performance of different representations. ``Trans" means the Transformer model. Mean and standard deviation of AUC or RMSE values are reported. For different metrics, "$\uparrow$" means the higher the better, and "$\downarrow$" contrarily. Results in bold and cells colored gray denote the best and the second best respectively.}
\centering
\begin{tabular}{lcccc|ccc}
\toprule
Task & \multicolumn{4}{c|}{Classification (ROC-AUC$\uparrow$ Higher is better)} & \multicolumn{3}{c}{Regression  (RMSE$\downarrow$ Lower is better)}\\
\midrule
Dataset &  BBBP & Tox21 & SIDER & ClinTox & FreeSolv & ESOL & Lipophilicity\\
\midrule
Trans&0.944 $\pm$ 0.011&0.813 $\pm$ 0.013&0.602 $\pm$ 0.012&0.954 $\pm$ 0.003&1.021 $\pm$ 0.102&0.767 $\pm$ 0.079&0.909 $\pm$ 0.023\\
\midrule
BMC &0.970 $\pm$ 0.002&0.857 $\pm$ 0.004&{\cellcolor{mygray}{0.679 $\pm$ 0.009}}&0.943 $\pm$ 0.030&0.776 $\pm$ 0.117&0.512 $\pm$ 0.025&{\cellcolor{mygray}{0.548 $\pm$ 0.006}}\\
\midrule
MMSG (w/o) &{\cellcolor{mygray}{0.973 $\pm$ 0.008}}&{\cellcolor{mygray}{0.858}} $\pm$ 0.003&0.677 $\pm$ 0.017&{\cellcolor{mygray}{0.979}} $\pm$ 0.009&{\cellcolor{mygray}{0.760}} $\pm$ 0.070&{\cellcolor{mygray}{0.511}} $\pm$ 0.029&0.551 $\pm$ 0.022\\
MMSG &\textbf{0.980 $\pm$ 0.004}&{\textbf{0.861 $\pm$ 0.031}}&\textbf{0.691 $\pm$ 0.015}&\textbf{0.993 $\pm$ 0.004}&\textbf{0.712 $\pm$ 0.087}&\textbf{0.495 $\pm$ 0.017}&\textbf{0.538 $\pm$ 0.007}\\
\bottomrule
\end{tabular}
\label{table 1}
\end{table*}

\subsection{Experimental Settings}
To comprehensively evaluate the learning ability of our MMSG, we follow MoleculeNet \cite{wu2018moleculenet} and split all the datasets by random splitting and scaffold splitting. The random splitting separates molecules in datasets randomly. The scaffold splitting, which is a more challenging and realistic setting, aims to separate molecules with different two-dimensional structural frameworks into different subsets. Both of the splitting ways are adopted with a ratio for train/validation/test sets as 0.8: 0.1: 0.1. We perform five independent runs with different random seeds on different splitting methods, and calculate the mean and standard deviation values of area under the receiver operating characteristic (ROC-AUC) or root-mean-squared error (RMSE) metric. The implementation of the model relies on Pytorch and RDKit package. TABLE 5 demonstrates the range of hyper-parameters used in experiments for MMSG, including the numbers of different network layers, batch size, learning rate, hidden dimension and so on.

\subsection{Results and Analysis}

\textbf{Quantitative results on the scaffold splitting}. TABLE 6 summarizes evaluations on scaffold splitting among parts of the baseline models, since some of the results are not reported previously. Cells in bold indicate the the best results on different datasets and the cells in gray rank second. TABLE 6 provides the following observations: 1) By combining SMILES and graph representations simultaneously, our MMSG framework achieves the state-of-the-art results on all benchmark datasets on scaffold splitting. The relatively improvement is 3.1$\%$ on classification tasks and 8.4$\%$ on regression tasks, and the total is 5.8$\%$ on all datasets. This fact proves the effectiveness of our multi-modal molecular joint representation learning framework with SMILES and graph. The results show that our MMSG can exactly extract a more comprehensive molecular representation for downstream tasks. 2) Comparing with GraSeq, our MMSG shows big enhancement by improving feature extraction modules of SMILES and graphs with considering the correspondence between them and the relatively improvement is 16.7$\%$ on average. 3) Our MMSG can obviously improve the property prediction performance especially on the small-scale datasets with only one task, like freeSolv, ESOL. This fact means that With combining multi-modal molecular representation, more hidden chemical information can be learned even if the data is limited, helping to improve the final molecular representation.

\textbf{Quantitative results on the random splitting}. TABLE 7 summarizes all the results on random splitting, and the definitions are same as TABLE 6. Due to the lack of results on scaffold splitting, we only collect the results of MPAD, HamNet, MoCL and GeomGCL on random splitting. Intuitively, our MMSG framework still outperforms most of the datasets, and our model only ranks second on Tox21 dataset, because the HamNet preserves 3D conformations of molecules that may have more affects on the toxicity. This fact also proves that 3D information of molecules is also deserved be considered for better multi-modal molecular representation since SMILES and graphs are both 2D molecular features. 

From both the results on scaffold splitting and random splitting, it can be obviously found that our MMSG shows significant performance on Clintox dataset. Similarly, MPAD also performs quite well on Clintox. An intuitive finding is that it may be easier to capture related chemical information from SMILES to make corresponding prediction tasks in ClinTox. This fact further proves the potential of learning molecular representation from multi-modal representations.

\subsection{Ablation studies}
To further investigate the positive influence of the modules proposed, several ablation experiments are conducted.

\textbf{The performance of different molecular representations}. In this study, we testify the potential value of the different molecular representations. Detailed results under random splitting are recorded in TABLE 8. ``Trans'' means the Transformer model which only uses SMILES representation as the input, while ``BMC'' means our BMC GNN which only uses graph as the input. ``MMSG (w/o)" means our MMSG combines both of the molecular representations without adding attention bias. TABLE 8 also shows that when only using singe-modal representation, SMILES-based model still performs better than graph-based model on Clintox dataset. This fact furthue proves the fact that using proper molecular representation is helpful to learn specific molecular properties. According to the results of ``MMSG (w/o)", with simply combing SMILES and graphs, the improvements are not attractive on all the tasks. With considering the chemical information between multi-modal representation, SMILES and graph in this work, more accurate molecular representation can be learned for various downstream task and the final results are also improved. 

\textbf{The performance of the bidirectional message communication GNN}. For verifying that our BMC GNN can exactly promote the information flow for better molecular representation of nodes and edges, we compare its performance with the original CPMNN under random splitting. Results are shown in TABLE 8, and the results re-confirm the effectiveness of our BMC GNN.

\begin{table}[t]
\caption{Ablation results of BMC GNN and CMPNN on seven datasets under random splitting, and results in cells bolded denote the best.}
\centering
\begin{tabular}{lcc}
\toprule
Dataset &  CMPNN &  BMC \\
\midrule
BBBP&0.963$\pm$0.003&\textbf{0.970$\pm$0.002}\\
Sider&0.666$\pm$0.007&\textbf{0.679$\pm$0.009}\\
Clintox&0.933$\pm$0.012&\textbf{0.943$\pm$0.030}\\
Tox21&0.856$\pm$0.007&\textbf{0.857$\pm$0.004}\\
\midrule
FreeSolv&0.819$\pm$0.147&\textbf{0.776$\pm$0.117}\\
ESOL&0.547$\pm$0.011&\textbf{0.512$\pm$0.025}\\
Lipophilicity&0.560$\pm$0.047&\textbf{0.548$\pm$0.006}\\
\bottomrule
\end{tabular}
\label{table 4}
\end{table}

\begin{figure*}[p]
\centering
\subfigure
[Transformer] 
{\includegraphics[width=5.6cm,height=5cm]{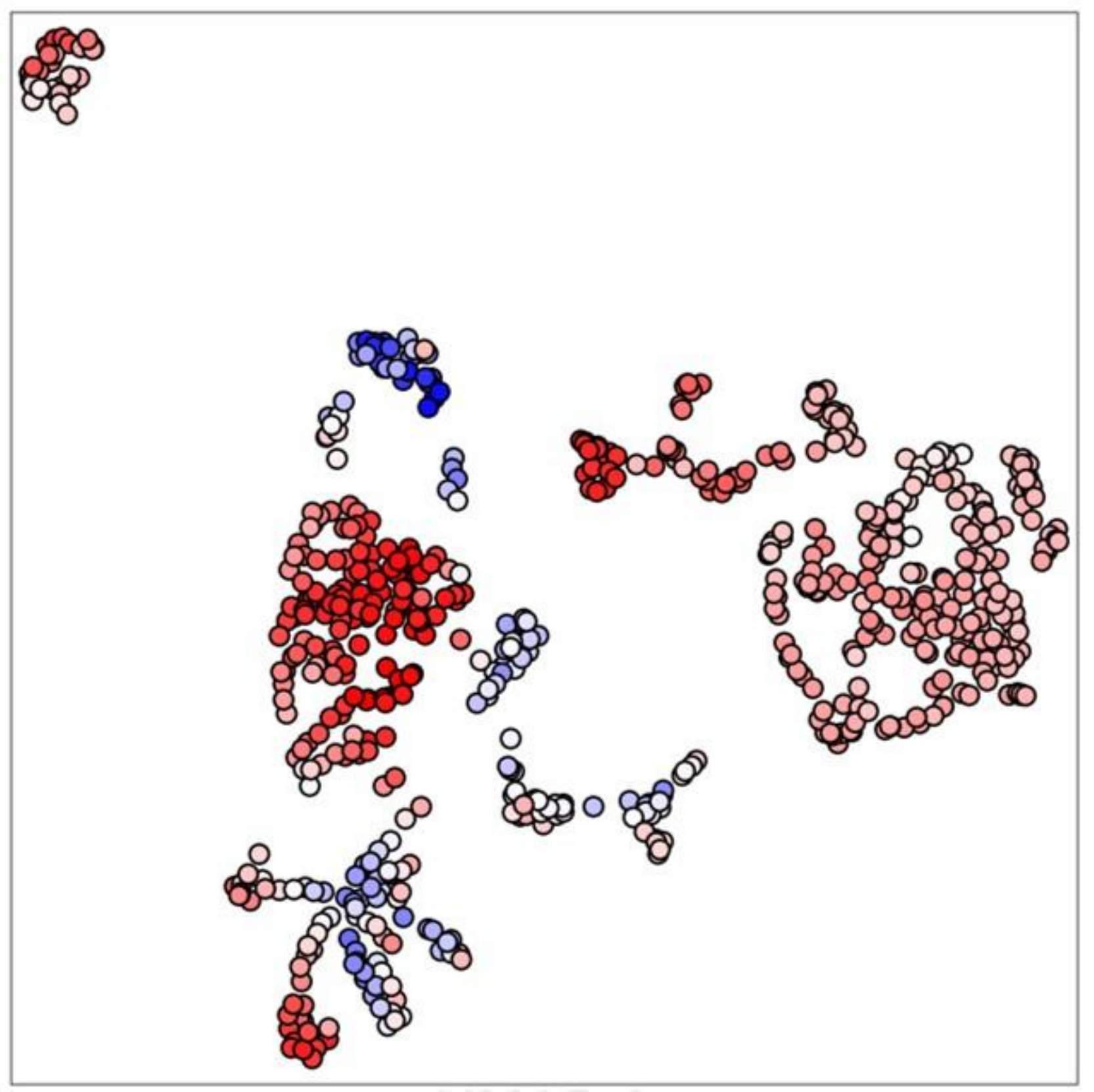}} 
\subfigure[CMPNN]{\includegraphics[width=5.6cm,height=5cm]{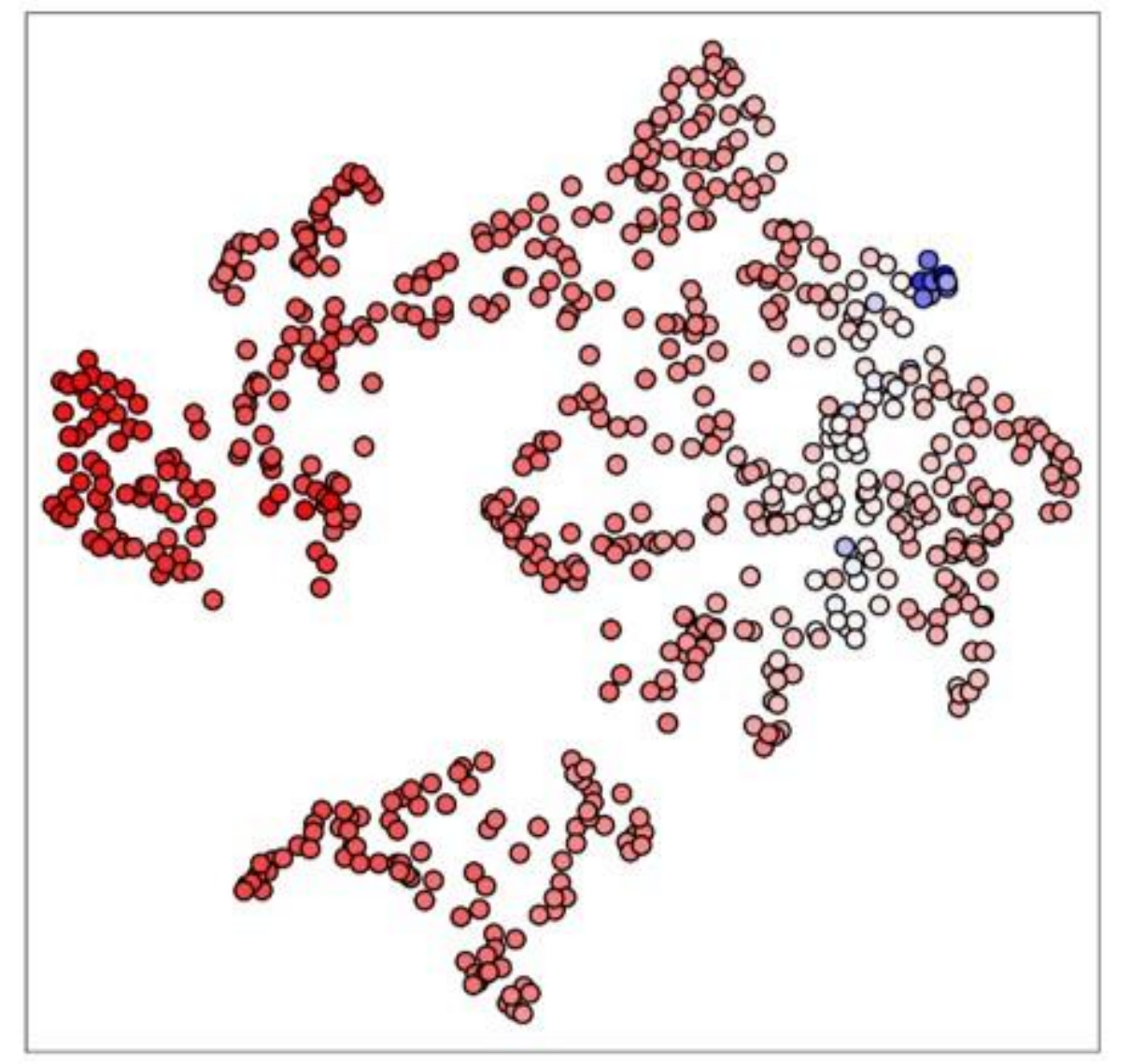}}
\subfigure[MMSG]{\includegraphics[width=5.6cm,height=5cm]{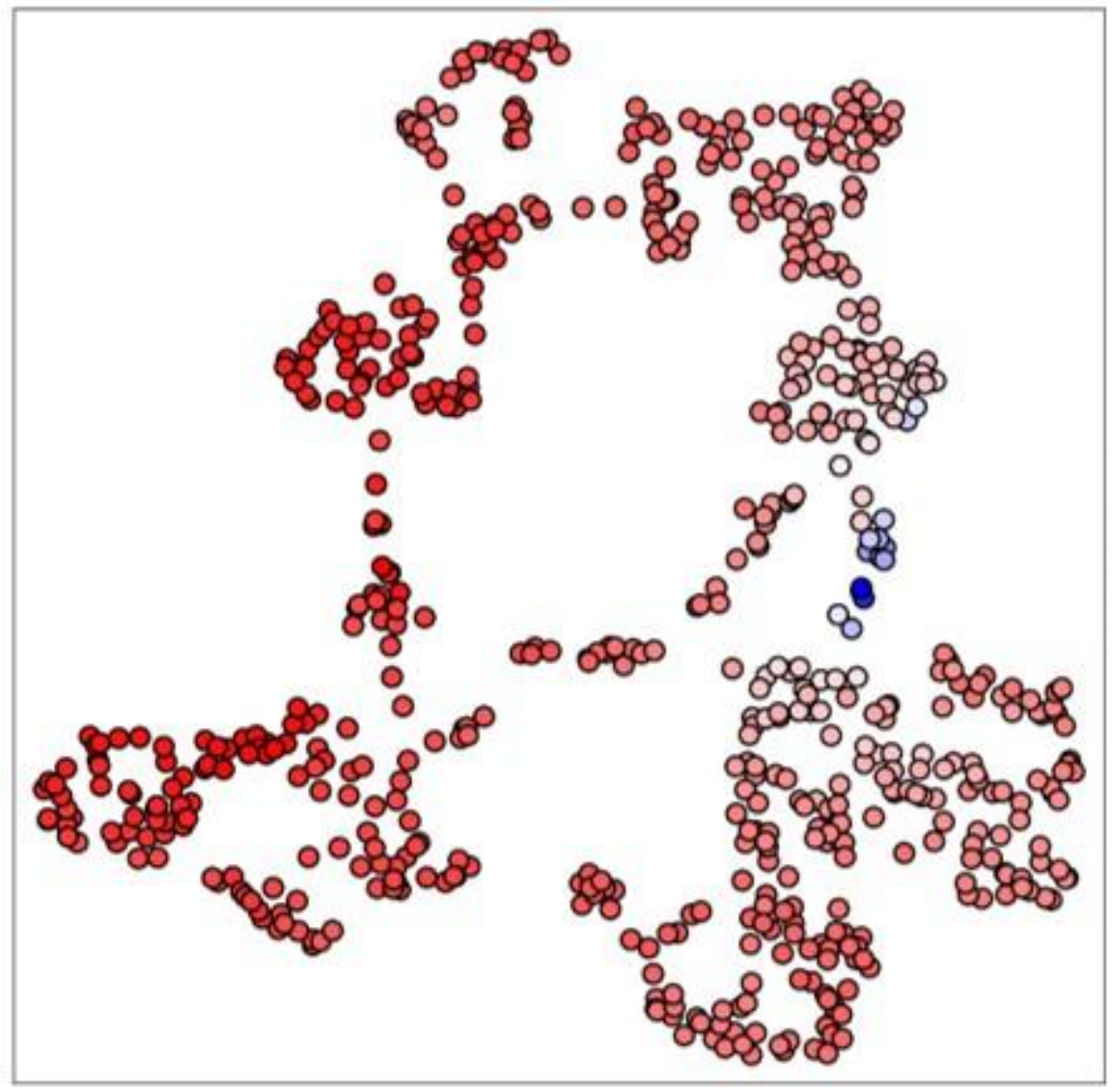}}
\subfigure{\includegraphics[height=5cm]{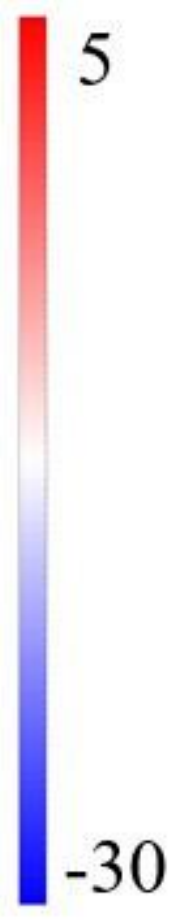}}
\caption{Visualization of the latent space by UMAP. Molecules are colored with the predicted values on the FreeSolv dataset by Transformer, CMPNN and MMSG. The color bar expresses the correspondence bewteen the value and the color.} 
\label{fig:1}  
\end{figure*}



\begin{figure*}[p]
\centering
\subfigure
[Transformer] 
{\includegraphics[width=5.6cm,height=5cm]{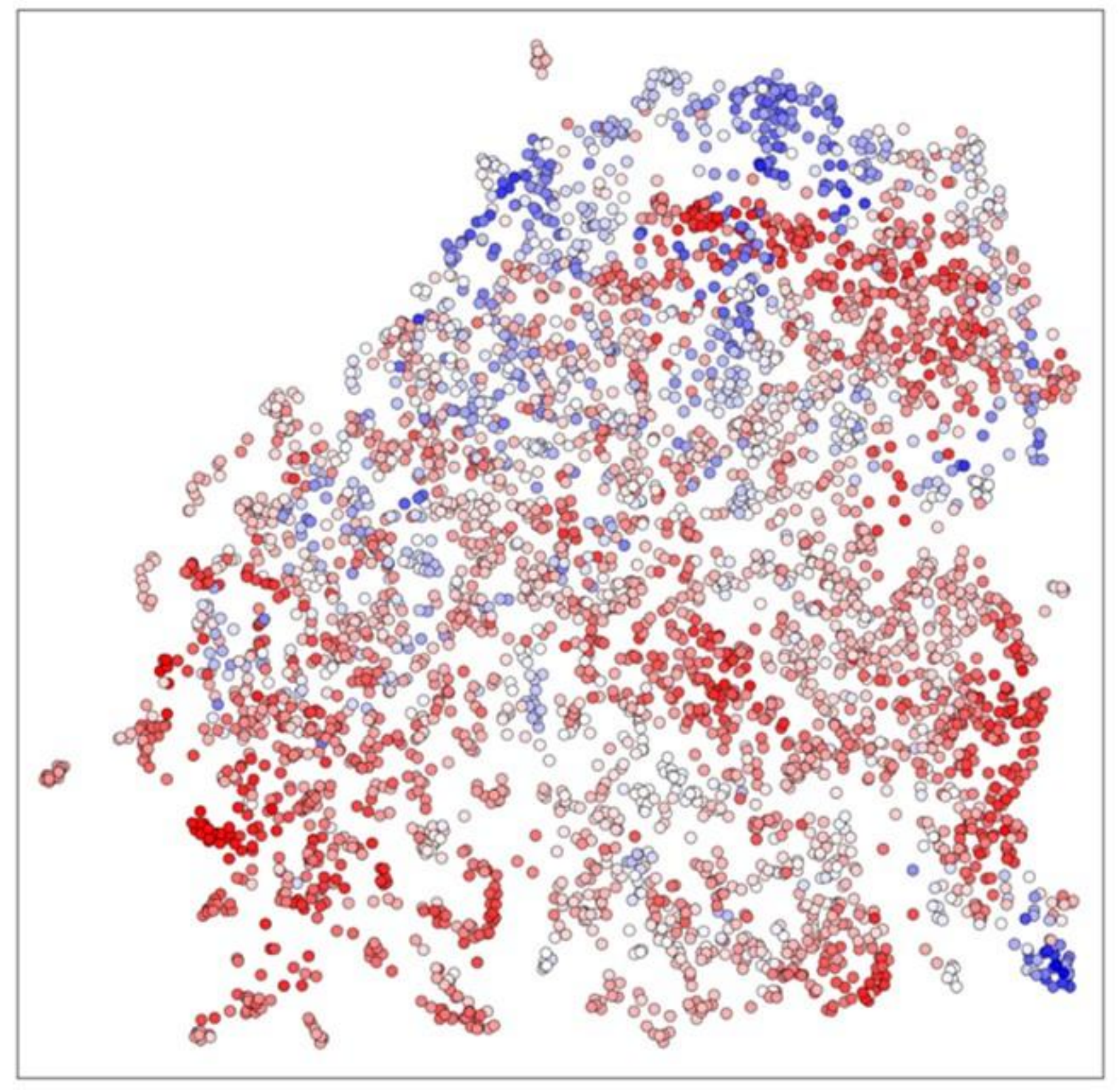}} 
\subfigure[CMPNN]{\includegraphics[width=5.6cm,height=5cm]{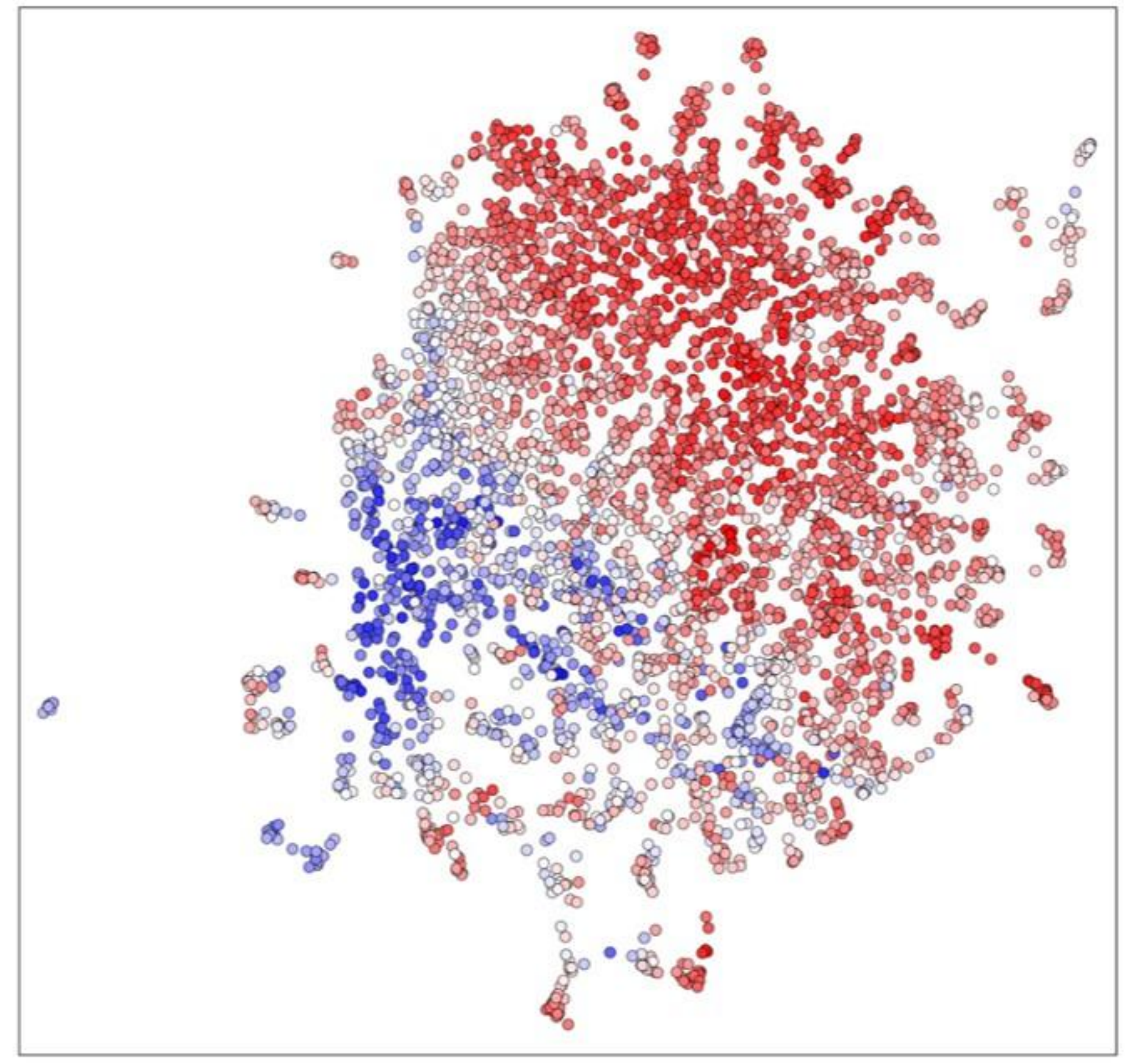}}
\subfigure[MMSG]{\includegraphics[width=5.6cm,height=5cm]{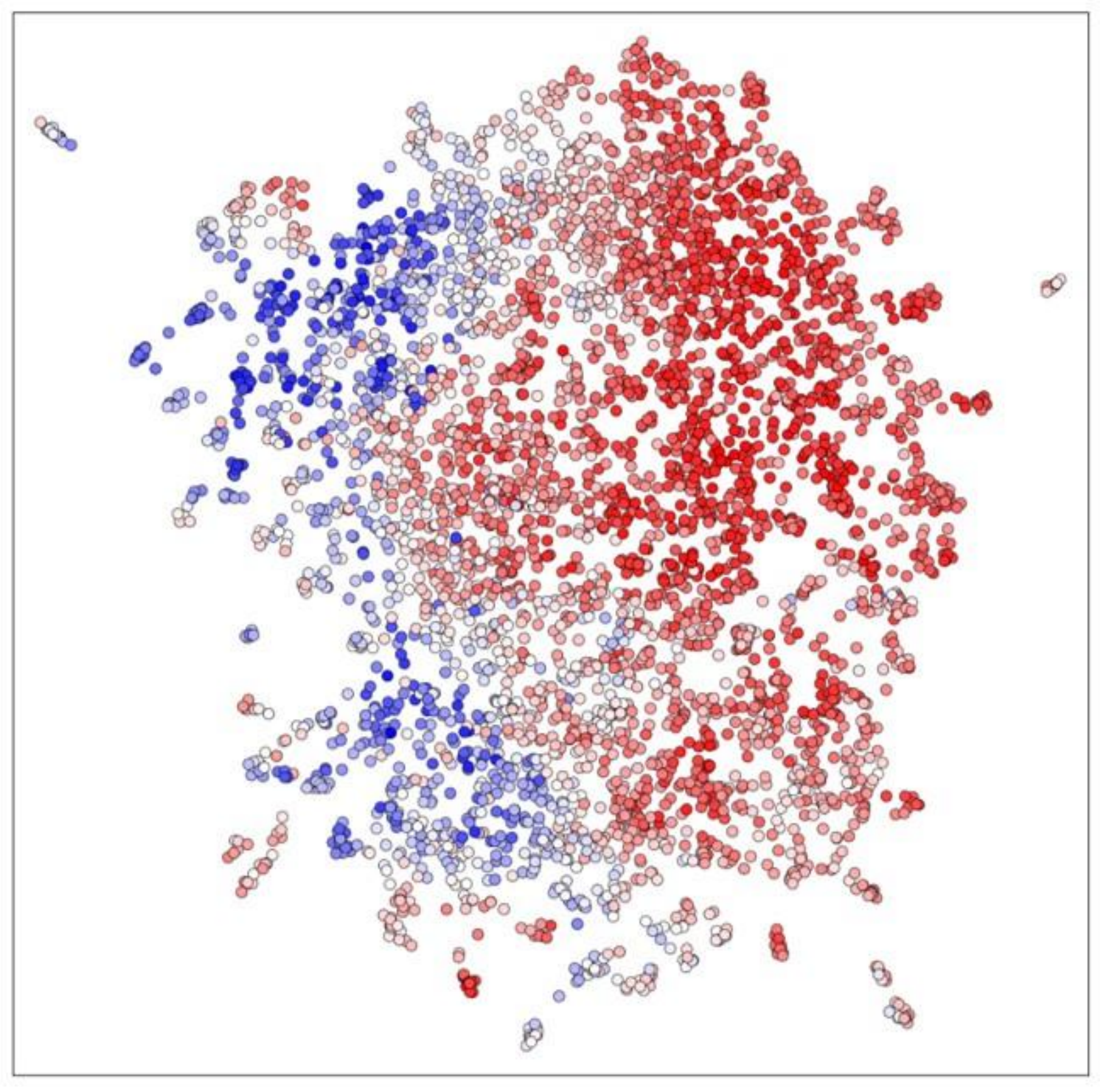}}
\subfigure{\includegraphics[height=5cm]{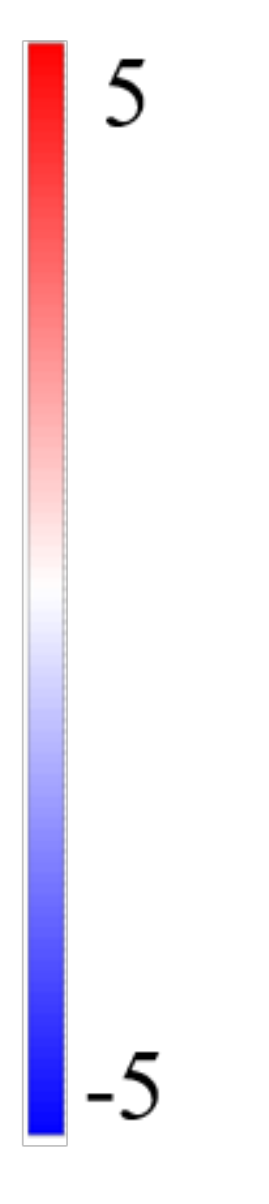}}
\caption{Visualization of the latent space by UMAP. Molecules are colored with the predicted values on the Lipophilicity dataset by Transformer, CMPNN and MMSG. The color bar expresses the correspondence bewteen the value and the color.} 
\label{fig:1}  
\end{figure*}

\begin{figure*}[p]
\centering
\subfigure
[Transformer] 
{\includegraphics[width=5.8cm,height=5.36cm]{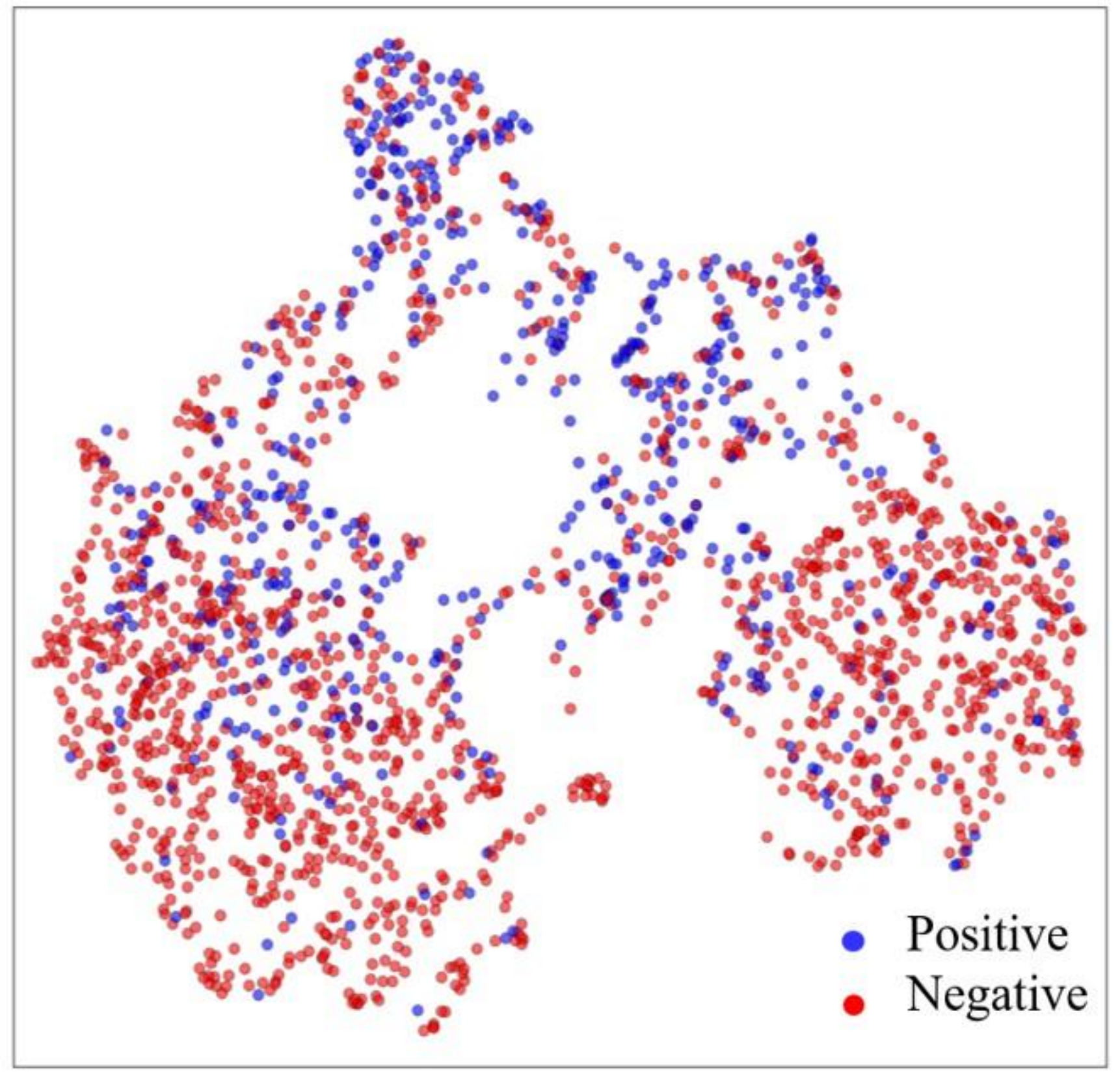}} 
\subfigure[CMPNN]{\includegraphics[width=5.8cm,height=5.36cm]{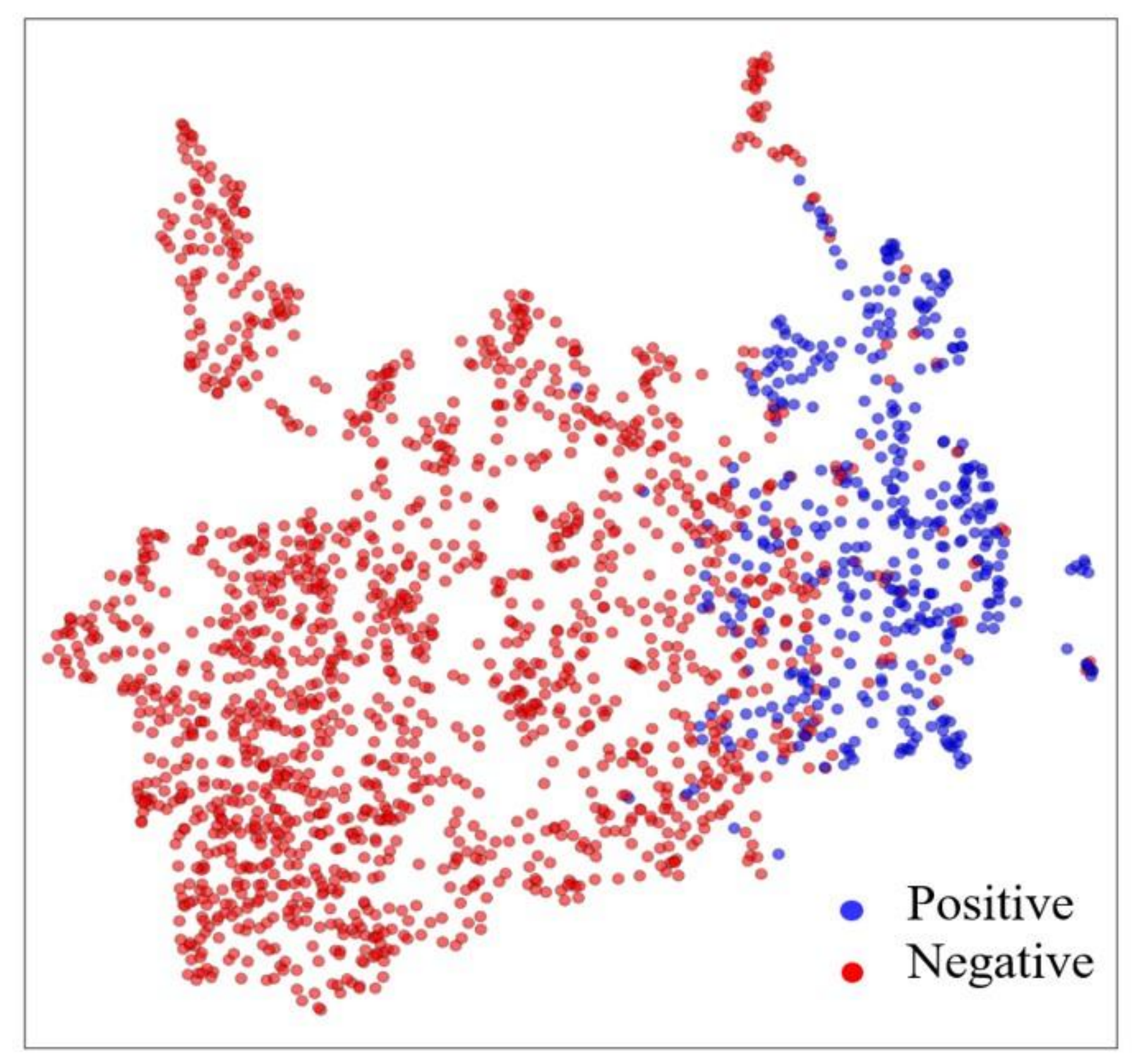}}
\subfigure[MMSG]{\includegraphics[width=5.8cm,height=5.36cm]{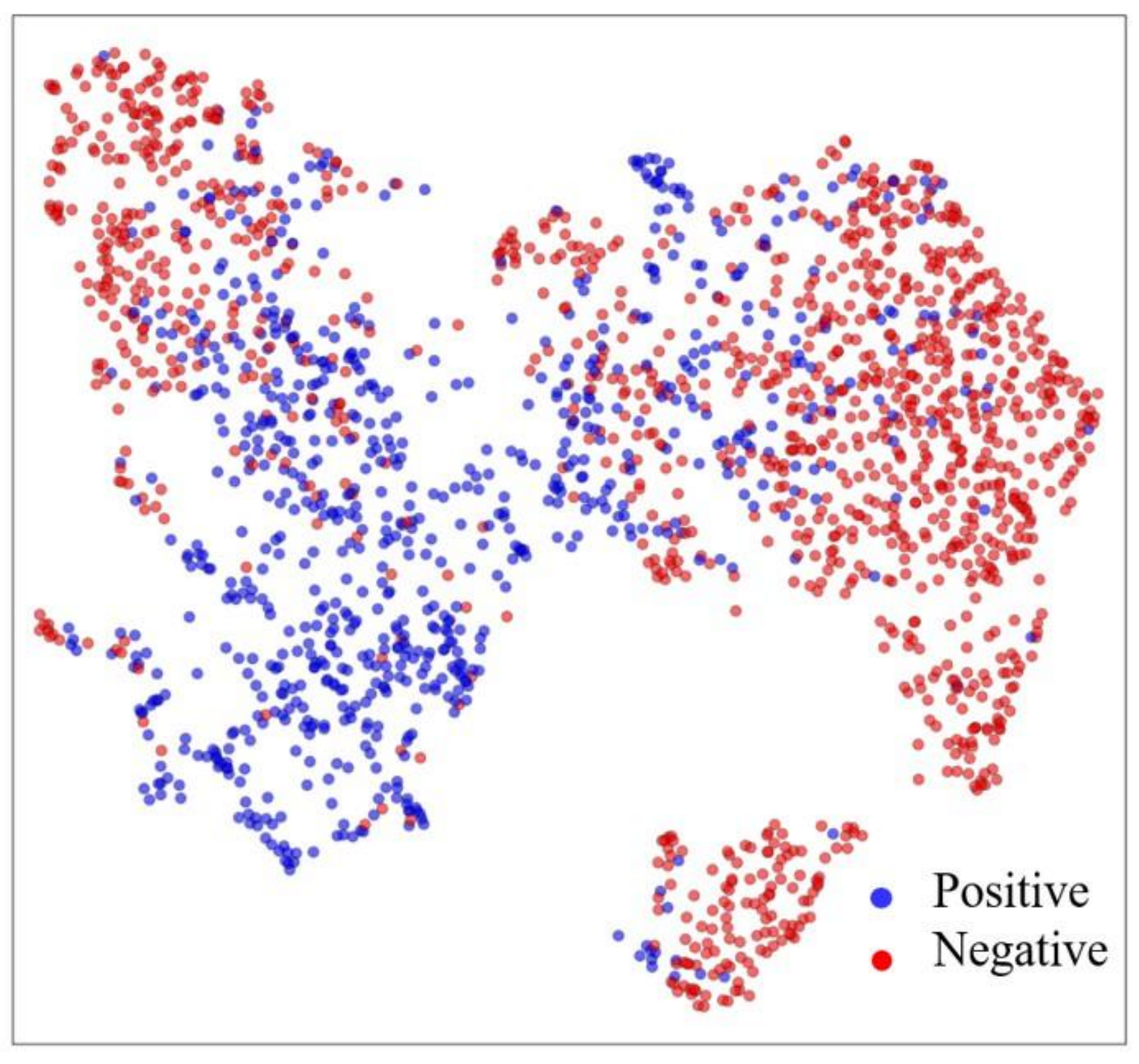}}
\caption{Visualization of the latent space by UMAP. Molecules are colored with the predicted results on the BBBP dataset by Transformer, CMPNN and MMSG. Blue points mean positive results and red points mean negative results.} 
\label{fig:1}  
\end{figure*}

\subsection{Molecular Representation Visualization}
As mentioned before, most of the GNN models face the problem of losing long-range dependencies which restricts the performance for molecular representation, and a case study is shown in Fig. 1. For molecules with similar structure, GNN model may project them into similar representations which are adjacent in corresponding chemical functional space, causing problematic final predictions. In our work, by realizing the chemical information correspondence between SMILES and graphs, our MMSG framework gains a more comprehensive molecular representation as expected. To prove this, we visualize the latent space by reducing their dimensions to 2 with UMAP. Projections results on FreeSolv, Lipophilicity and BBBP are illustrated in Fig. 4, Fig. 5 and Fig. 6. Each point corresponds to a molecule and is colored according to its predicted value.

For small-scale dataset FreeSolv (Fig. 4), all the sub-figures show reasonable results. Generally, molecules with similar physical properties are usually clustered nearly in their chemical space. Transformer model only uses the SMILES encoding which generally catch the global information of molecules, so most of the points are clustered in (a), while this distribution does not correspond to the facts. In contrast, both the CMPNN (b) and our MMSG (c) distribute reasonably. There exists clear gradation from upper and lower left to middle right. Moreover, our MMSG framework shows more dense clusters than CMPNN. 

For large-scale dataset Lipophilicity (Fig.5) and BBBP (Fig 6.), we can get the same results. From Fig 5.(b) and Fig 6.(b), it can be clearly found that molecules with positive properties and molecules have clear boundaries. While in Fig 5.(c) and Fig 6.(c), molecules show more delicate difference instead clustering together. In the way, these results suggests that our MMSG model capture better representations of molecules. 
\section{Conclusion}

In this paper, we propose a novel molecular joint representation learning framework, called MMSG, via multi-modal molecular information (from SMILES and graphs). We further consider the correspondence between chemical information embedded in multi-modal representations for completely different encoding rules. We modify the self-attention module in Transformer to reinforce the bond-level information correspondence between SMILES and graphs. Further, we realize bidirectional communication between nodes and edges in directed graphs so as to improve the information flow. Numerous experiments on different public datasets prove the effectiveness of our proposed model, with surpassing state-of-the-art results. 

Moreover, in this work, we only consider the usage of SMILES and molecular graphs, while our framework has strong portability. More means of molecular representations, e.g. 3D structure of a molecule, can be taken into consideration to gain specific chemical information. More research will be investigated in future work.

\bibliographystyle{IEEEtran}
\bibliography{TCBB.bib}

\begin{thebibliography}{10}
\providecommand{\url}[1]{#1}
\csname url@samestyle\endcsname
\providecommand{\newblock}{\relax}
\providecommand{\bibinfo}[2]{#2}
\providecommand{\BIBentrySTDinterwordspacing}{\spaceskip=0pt\relax}
\providecommand{\BIBentryALTinterwordstretchfactor}{4}
\providecommand{\BIBentryALTinterwordspacing}{\spaceskip=\fontdimen2\font plus
\BIBentryALTinterwordstretchfactor\fontdimen3\font minus
  \fontdimen4\font\relax}
\providecommand{\BIBforeignlanguage}[2]{{%
\expandafter\ifx\csname l@#1\endcsname\relax
\typeout{** WARNING: IEEEtran.bst: No hyphenation pattern has been}%
\typeout{** loaded for the language `#1'. Using the pattern for}%
\typeout{** the default language instead.}%
\else
\language=\csname l@#1\endcsname
\fi
#2}}
\providecommand{\BIBdecl}{\relax}
\BIBdecl

\bibitem{dimasi2016innovation}
J.~A. DiMasi, H.~G. Grabowski, and R.~W. Hansen, ``Innovation in the
  pharmaceutical industry: new estimates of r\&d costs,'' \emph{Journal of
  Health Economics}, vol.~47, pp. 20--33, 2016.

\bibitem{9834316}
Y.~Qian, Y.~Ding, Q.~Zou, and F.~Guo, ``Multi-view kernel sparse representation
  for identification of membrane protein types,'' \emph{IEEE/ACM Transactions
  on Computational Biology and Bioinformatics}, pp. 1--12, 2022.

\bibitem{8880542}
J.~Guan, R.~Li, S.~Yu, and X.~Zhang, ``A method for generating synthetic
  electronic medical record text,'' \emph{IEEE/ACM Transactions on
  Computational Biology and Bioinformatics}, vol.~18, no.~1, pp. 173--182,
  2021.

\bibitem{jumper2021highly}
J.~Jumper, R.~Evans, A.~Pritzel, T.~Green, M.~Figurnov, O.~Ronneberger,
  K.~Tunyasuvunakool, R.~Bates, A.~{\v{Z}}{\'\i}dek, A.~Potapenko
  \emph{et~al.}, ``Highly accurate protein structure prediction with
  alphafold,'' \emph{Nature}, pp. 1--11, 2021.

\bibitem{jimenez2020drug}
J.~Jim{\'e}nez-Luna, F.~Grisoni, and G.~Schneider, ``Drug discovery with
  explainable artificial intelligence,'' \emph{Nature Machine Intelligence},
  vol.~2, no.~10, pp. 573--584, 2020.

\bibitem{9387544}
Q.~Zhao, M.~Yang, Z.~Cheng, Y.~Li, and J.~Wang, ``Biomedical data and deep
  learning computational models for predicting compound-protein relations,''
  \emph{IEEE/ACM Transactions on Computational Biology and Bioinformatics},
  vol.~19, no.~4, pp. 2092--2110, 2022.

\bibitem{rogers2010extended}
D.~Rogers and M.~Hahn, ``Extended-connectivity fingerprints,'' \emph{Journal of
  chemical information and modeling}, vol.~50, no.~5, pp. 742--754, 2010.

\bibitem{tang2020self}
B.~Tang, S.~T. Kramer, M.~Fang, Y.~Qiu, Z.~Wu, and D.~Xu, ``A self-attention
  based message passing neural network for predicting molecular lipophilicity
  and aqueous solubility,'' \emph{Journal of Cheminformatics}, vol.~12, no.~1,
  pp. 1--9, 2020.

\bibitem{guo2020graseq}
Z.~Guo, W.~Yu, C.~Zhang, M.~Jiang, and N.~V. Chawla, ``Graseq: graph and
  sequence fusion learning for molecular property prediction,'' in
  \emph{Proceedings of the Twenty-ninth ACM International Conference on
  Information \& Knowledge Management (CIKM)}, Online, 2020, pp. 435--443.

\bibitem{weininger1988smiles}
D.~Weininger, ``Smiles, a chemical language and information system. 1.
  introduction to methodology and encoding rules,'' \emph{Journal of Chemical
  Information and Computer Sciences}, vol.~28, no.~1, pp. 31--36, 1988.

\bibitem{vaswani2017attention}
A.~Vaswani, N.~Shazeer, N.~Parmar, J.~Uszkoreit, L.~Jones, A.~N. Gomez,
  {\L}.~Kaiser, and I.~Polosukhin, ``Attention is all you need,'' in
  \emph{Proceedings of the Thirty-first Conference on Neural Information
  Processing Systems (NIPS)}, Long Beach, USA, 2017, pp. 5998--6008.

\bibitem{9866850}
Y.~Wan and Z.~Jiang, ``Transcrispr: Transformer based hybrid model for
  predicting crispr/cas9 single guide rna cleavage efficiency,'' \emph{IEEE/ACM
  Transactions on Computational Biology and Bioinformatics}, pp. 1--12, 2022.

\bibitem{song2020communicative}
Y.~Song, S.~Zheng, Z.~Niu, Z.-H. Fu, Y.~Lu, and Y.~Yang, ``Communicative
  representation learning on attributed molecular graphs,'' in
  \emph{Proceedings of the Twenty-ninth International Joint Conference on
  Artificial Intelligence (IJCAI)}, Yokoham, Japan, 2020, pp. 2831--2838.

\bibitem{9864145}
J.~Chen, J.~Gao, T.~Lyu, B.~M. Oloulade, and X.~Hu, ``Automsr: Auto molecular
  structure representation learning for multi-label metabolic pathway
  prediction,'' \emph{IEEE/ACM Transactions on Computational Biology and
  Bioinformatics}, pp. 1--11, 2022.

\bibitem{sun2021mocl}
M.~Sun, J.~Xing, H.~Wang, B.~Chen, and J.~Zhou, ``Mocl: data-driven molecular
  fingerprint via knowledge-aware contrastive learning from molecular graph,''
  in \emph{Proceedings of the Twenty-seventh ACM SIGKDD Conference on Knowledge
  Discovery \& Data Mining}, Online, 2021, pp. 3585--3594.

\bibitem{li2022geomgcl}
S.~Li, J.~Zhou, T.~Xu, D.~Dou, and H.~Xiong, ``Geomgcl: Geometric graph
  contrastive learning for molecular property prediction,'' \emph{arXiv
  preprint arXiv:2109.11730}, 2021.

\bibitem{zhu2020beyond}
J.~Zhu, Y.~Yan, L.~Zhao, M.~Heimann, L.~Akoglu, and D.~Koutra, ``Beyond
  homophily in graph neural networks: Current limitations and effective
  designs,'' \emph{Advances in Neural Information Processing Systems}, vol.~33,
  pp. 7793--7804, 2020.

\bibitem{mao2021molecular}
K.~Mao, X.~Xiao, T.~Xu, Y.~Rong, J.~Huang, and P.~Zhao, ``Molecular graph
  enhanced transformer for retrosynthesis prediction,'' \emph{Neurocomputing},
  vol. 457, pp. 193--202, 2021.

\bibitem{paul2018chemixnet}
A.~Paul, D.~Jha, R.~Al-Bahrani, W.-k. Liao, A.~Choudhary, and A.~Agrawal,
  ``Chemixnet: Mixed dnn architectures for predicting chemical properties using
  multiple molecular representations,'' \emph{arXiv preprint arXiv:1811.08283},
  2018.

\bibitem{ozturk2020exploring}
H.~{\"O}zt{\"u}rk, A.~{\"O}zg{\"u}r, P.~Schwaller, T.~Laino, and E.~Ozkirimli,
  ``Exploring chemical space using natural language processing methodologies
  for drug discovery,'' \emph{Drug Discovery Today}, vol.~25, no.~4, pp.
  689--705, 2020.

\bibitem{zheng2019identifying}
S.~Zheng, X.~Yan, Y.~Yang, and J.~Xu, ``Identifying structure--property
  relationships through smiles syntax analysis with self-attention mechanism,''
  \emph{Journal of Chemical Information and Modeling}, vol.~59, no.~2, pp.
  914--923, 2019.

\bibitem{honda2019smiles}
S.~Honda, S.~Shi, and H.~R. Ueda, ``Smiles transformer: Pre-trained molecular
  fingerprint for low data drug discovery,'' \emph{arXiv preprint
  arXiv:1911.04738}, 2019.

\bibitem{wang2019smiles}
S.~Wang, Y.~Guo, Y.~Wang, H.~Sun, and J.~Huang, ``Smiles-bert: large scale
  unsupervised pre-training for molecular property prediction,'' in \emph{The
  Tenth ACM Conference on Bioinformatics, Computational Biology, and Health
  Informatics (ACM BCB)}, NewYork, USA, 2019, pp. 429--436.

\bibitem{jo2020message}
J.~Jo, B.~Kwak, H.-S. Choi, and S.~Yoon, ``The message passing neural networks
  for chemical property prediction on smiles,'' \emph{Methods}, vol. 179, pp.
  65--72, 2020.

\bibitem{sun2020graph}
M.~Sun, S.~Zhao, C.~Gilvary, O.~Elemento, J.~Zhou, and F.~Wang, ``Graph
  convolutional networks for computational drug development and discovery,''
  \emph{Briefings in Bioinformatics}, vol.~21, no.~3, pp. 919--935, 2020.

\bibitem{hamilton2017inductive}
W.~L. Hamilton, R.~Ying, and J.~Leskovec, ``Inductive representation learning
  on large graphs,'' in \emph{Proceedings of the Thirty-first Conference on
  Neural Information Processing Systems (NIPS)}, Long Beach, USA, 2017, pp.
  1025--1035.

\bibitem{xu2018powerful}
K.~Xu, W.~Hu, J.~Leskovec, and S.~Jegelka, ``How powerful are graph neural
  networks?'' in \emph{Proceedings of the Sixth International Conference on
  Learning Representations (ICLR)}, Vancouver, Canada, 2018.

\bibitem{gilmer2017neural}
J.~Gilmer, S.~S. Schoenholz, P.~F. Riley, O.~Vinyals, and G.~E. Dahl, ``Neural
  message passing for quantum chemistry,'' in \emph{Proceedings of the
  Thirty-fourth International Conference on Machine Learning (ICML)}, Sydney,
  Australia, 2017, pp. 1263--1272.

\bibitem{yang2019analyzing}
K.~Yang, K.~Swanson, W.~Jin, C.~Coley, P.~Eiden, H.~Gao, A.~Guzman-Perez,
  T.~Hopper, B.~Kelley, M.~Mathea \emph{et~al.}, ``Analyzing learned molecular
  representations for property prediction,'' \emph{Journal of Chemical
  Information and Modeling}, vol.~59, no.~8, pp. 3370--3388, 2019.

\bibitem{zhang2019gresnet}
J.~Zhang and L.~Meng, ``Gresnet: Graph residual network for reviving deep gnns
  from suspended animation,'' \emph{arXiv preprint arXiv:1909.05729}, 2019.

\bibitem{rong2020self}
Y.~Rong, Y.~Bian, T.~Xu, W.~Xie, Y.~Wei, W.~Huang, and J.~Huang,
  ``Self-supervised graph transformer on large-scale molecular data,'' in
  \emph{Proceedings of the Thirty-fourth Conference on Neural Information
  Processing Systems (NIPS)}, Online, 2020, pp. 12\,559--12\,571.

\bibitem{chen2021learning}
J.~Chen, S.~Zheng, Y.~Song, J.~Rao, and Y.~Yang, ``Learning attributed graph
  representations with communicative message passing transformer,'' in
  \emph{Proceedings of the Thirtieth International Joint Conference on
  Artificial Intelligence (IJCAI)}, Montréal, Canada, 2021.

\bibitem{kipf2016semi}
T.~N. Kipf and M.~Welling, ``Semi-supervised classification with graph
  convolutional networks,'' in \emph{Proceedings of the Fifth International
  Conference on Learning Representations (ICLR)}, Toulon, France, 2017.

\bibitem{chung2014empirical}
J.~Chung, C.~Gulcehre, K.~Cho, and Y.~Bengio, ``Empirical evaluation of gated
  recurrent neural networks on sequence modeling,'' in \emph{Proceedings of the
  Twenty-eighth Conference on Neural Information Processing Systems (NIPS)
  Workshop on Deep Learning}, Montréal, Canada, 2014.

\bibitem{9425008}
Z.~Cheng, C.~Yan, F.-X. Wu, and J.~Wang, ``Drug-target interaction prediction
  using multi-head self-attention and graph attention network,'' \emph{IEEE/ACM
  Transactions on Computational Biology and Bioinformatics}, vol.~19, no.~4,
  pp. 2208--2218, 2022.

\bibitem{9456970}
Y.~Guo, O.~Krupa, J.~Stein, G.~Wu, and A.~Krishnamurthy, ``Sau-net: A unified
  network for cell counting in 2d and 3d microscopy images,'' \emph{IEEE/ACM
  Transactions on Computational Biology and Bioinformatics}, vol.~19, no.~4,
  pp. 1920--1932, 2022.

\bibitem{wu2018moleculenet}
Z.~Wu, B.~Ramsundar, E.~N. Feinberg, J.~Gomes, C.~Geniesse, A.~S. Pappu,
  K.~Leswing, and V.~Pande, ``Moleculenet: a benchmark for molecular machine
  learning,'' \emph{Chemical Science}, vol.~9, no.~2, pp. 513--530, 2018.

\bibitem{martins2012bayesian}
I.~F. Martins, A.~L. Teixeira, L.~Pinheiro, and A.~O. Falcao, ``A bayesian
  approach to in silico blood-brain barrier penetration modeling,''
  \emph{Journal of Chemical Information and Modeling}, vol.~52, no.~6, pp.
  1686--1697, 2012.

\bibitem{tox212017}
{NCATS}, ``Tox21 challenge,'' 2017, accessed: 2017-09-07.

\bibitem{kuhn2016sider}
M.~Kuhn, I.~Letunic, L.~J. Jensen, and P.~Bork, ``The sider database of drugs
  and side effects,'' \emph{Nucleic Acids Research}, vol.~44, no.~D1, pp.
  D1075--D1079, 2016.

\bibitem{gayvert2016data}
K.~M. Gayvert, N.~S. Madhukar, and O.~Elemento, ``A data-driven approach to
  predicting successes and failures of clinical trials,'' \emph{Cell Chemical
  Biology}, vol.~23, no.~10, pp. 1294--1301, 2016.

\bibitem{mobley2014freesolv}
D.~L. Mobley and J.~P. Guthrie, ``Freesolv: a database of experimental and
  calculated hydration free energies, with input files,'' \emph{Journal of
  Computer-aided Molecular Design}, vol.~28, no.~7, pp. 711--720, 2014.

\bibitem{delaney2004esol}
J.~S. Delaney, ``Esol: estimating aqueous solubility directly from molecular
  structure,'' \emph{Journal of chemical information and computer sciences},
  vol.~44, no.~3, pp. 1000--1005, 2004.

\bibitem{wenlock2015experimental}
M.~Wenlock and N.~Tomkinson, ``Experimental in vitro dmpk and physicochemical
  data on a set of publicly disclosed compounds,'' 2015.

\bibitem{kearnes2016molecular}
S.~Kearnes, K.~McCloskey, M.~Berndl, V.~Pande, and P.~Riley, ``Molecular graph
  convolutions: moving beyond fingerprints,'' \emph{Journal of Computer-aided
  Molecular Design}, vol.~30, no.~8, pp. 595--608, 2016.

\bibitem{liu2019n}
S.~Liu, M.~F. Demirel, and Y.~Liang, ``N-gram graph: simple unsupervised
  representation for graphs, with applications to molecules,'' in
  \emph{Proceedings of the Thirty-third Conference on Neural Information
  Processing Systems (NIPS)}, Vancouver, Canada, 2019, pp. 8466--8478.

\bibitem{xiong2019pushing}
Z.~Xiong, D.~Wang, X.~Liu, F.~Zhong, X.~Wan, X.~Li, Z.~Li, X.~Luo, K.~Chen,
  H.~Jiang \emph{et~al.}, ``Pushing the boundaries of molecular representation
  for drug discovery with the graph attention mechanism,'' \emph{Journal of
  Medicinal Chemistry}, vol.~63, no.~16, pp. 8749--8760, 2019.

\bibitem{li2021hamnet}
Z.~Li, S.~Yang, G.~Song, and L.~Cai, ``Hamnet: Conformation-guided molecular
  representation with hamiltonian neural networks,'' in \emph{Proceedings of
  the Tenth International Conference on Learning Representations (ICLR)},
  Online, 2021.

\end{thebibliography}
%








\end{document}